\documentclass[10pt,twocolumn,letterpaper]{article}

\usepackage[pagenumbers]{cvpr}

\usepackage{pifont}

\usepackage{siunitx}
\usepackage{makecell}
\usepackage{graphicx}
\usepackage{multirow}
\usepackage{setspace}
\usepackage{dcolumn}
\usepackage{comment}
\usepackage{colortbl}
\usepackage{pifont}
\usepackage{placeins}

\usepackage[pagebackref,breaklinks,colorlinks]{hyperref}
\definecolor{darkgreen}{rgb}{0.0, 0.5, 0.0}
\newcommand{\mypar}[1]{\vspace{1mm}\noindent\textbf{#1}}

\def\method{\textsc{Find3D}\xspace}
\def\siglip{SigLIP\xspace}
\def\shapenetpart{ShapeNet-Part\xspace}
\def\objaversegeneral{Objaverse-General\xspace}

\def\objaverseunseen{Objaverse-Unseen Categories\xspace}
\def\shapenetpartobj{ShapeNetPart-V2\xspace}
\def\partnete{PartNet-E\xspace}
\def\check{\textcolor{darkgreen}{\checkmark}}
\def\cross{\textcolor{red}{\ding{55}}}

\title{Find Any Part in 3D}

\author{Ziqi Ma
\qquad
Yisong Yue
\qquad
Georgia Gkioxari
\\
California Institute of Technology
}

\begin{document}

\twocolumn[{%
\renewcommand\twocolumn[1][]{#1}%
\maketitle
\begin{center}
    \centering
    \includegraphics[width=\linewidth]{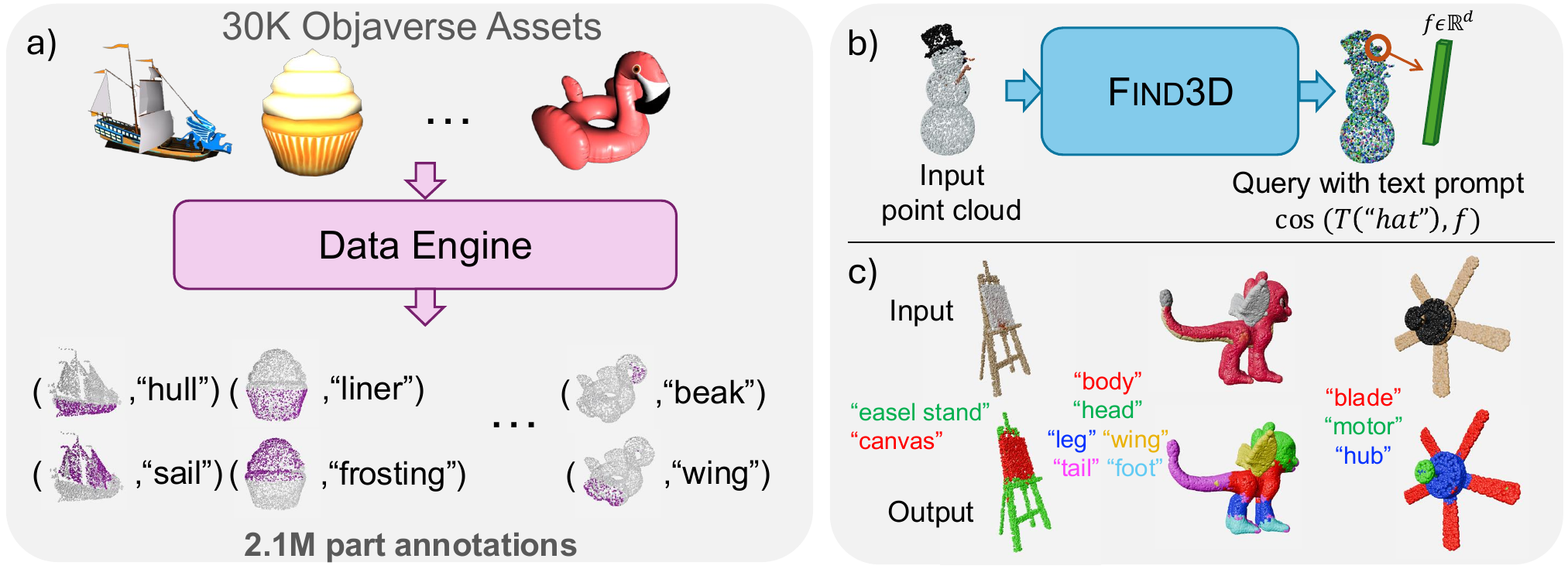}
    \vspace{-7mm}
    \captionof{figure}{\method is the first general-category 3D model that can segment \emph{any} part of \emph{any} object with \emph{any} text query. We achieve this by building a scalable \textbf{Data Engine} powered by 2D foundation models -- SAM \& Gemini -- that automatically annotates 3D assets from the web. Using the labeled data, \textbf{\method} trains a transformer-based point cloud model with a contrastive training recipe. Our method works on diverse 3D objects and parts, \eg the easel, the imaginary animal, and the ceiling fan.}
    \label{fig:schematic}
\end{center}
}]

\begin{abstract}
Why don't we have foundation models in 3D yet? A key limitation is data scarcity. For 3D object part segmentation, existing datasets are small in size and lack diversity. We show that it is possible to break this data barrier by building a data engine powered by 2D foundation models. Our data engine automatically annotates any number of object parts: \textbf{1755$\times$} more unique part types than existing datasets combined. By training on our annotated data with a simple contrastive objective, we obtain an open-world model that generalizes to \textbf{any} part in \textbf{any} object based on \textbf{any} text query. Even when evaluated zero-shot, we outperform existing methods on the datasets they train on. We achieve $\mathbf{260\%}$ \textbf{improvement in mIoU} and boost speed by \textbf{6$\times$} to \textbf{300$\times$}.
Our scaling analysis confirms that this generalization stems from the data scale, which underscores the impact of our data engine.
Finally, to advance general-category open-world 3D part segmentation, we release a benchmark covering a wide range of objects and parts.
Project website: \url{https://ziqi-ma.github.io/find3dsite/}
\end{abstract}    
\section{Introduction}
\label{sec:intro}

Is it possible to build foundation models in 3D? For text and image modalities, we have seen that strong, general models come from internet-scale training data. In the absence of such large-scale 3D datasets, can we attempt to replicate this success in 3D?

In this paper, we provide an answer to this question. We show that when you tackle the data challenge, you can get a strong model with a simple, general training recipe. This approach not only unlocks generalization to unseen objects with a $\mathbf{260\%}$ \textbf{improvement in mIoU}, but even outperforms prior methods on the datasets they train on.

Prior works, heavy on dataset-specific customization, suffer from poor generalization because existing datasets are small and homogeneous. For example, \shapenetpart~\cite{yi2016scalable} contains only 16 categories and makes assumptions such as ``all chairs face right". Evaluation on such limited datasets implicitly encourages dataset-specific customization, which is not the path towards generalization. Moreover, many prior works use pipelines that are computationally expensive and cannot scale to larger training sets.

In this paper, our goal is to: 1) establish a scalable data engine that can generate useful labels for any number of 3D assets; and 2) show that having a large-scale training set enables strong generalization with a simple training recipe, without any customizations for specific datasets such as per-category prompt and viewpoint search~\cite{zhu2023pointclip}, category-specific finetuning~\cite{zhou2023partslip++, umam2024partdistill}, multi-pass inference customization with predefined part ranking logic~\cite{zhou2023partslip++}, or slow test-time pipelines~\cite{takmaz2023openmask3d}. Conceptually, our findings mirror those in other domains such as text, where using general training recipes at scale leads to powerful and general foundation models.

Concretely, as shown in \cref{fig:schematic}, we enable scaling in 3D by building a data engine that automatically annotates synthetic 3D assets on the internet, yielding $\mathbf{2.1}$ \textbf{million} part annotations of 761 object categories. Our dataset contains $124615$ unique part types, which is over $\mathbf{1775\times}$ the number of unique part types in existing datasets combined (\shapenetpart~\cite{yi2016scalable} and \partnete~\cite{mo2019partnet} contain 71 unique part types combined). To leverage such large-scale data, we devise a contrastive training objective to handle part hierarchy and ambiguity. Our model takes in a point cloud and predicts a queryable semantic feature for every point. The features are in the latent embedding space of a CLIP-like~\cite{radford2021learning} model, so that they can be queried with any free-form text by calculating pointwise cosine similarities with the query embedding.

This approach yields a model that can segment \emph{any} part of \emph{any} object, with \emph{any} text query. We highlight the following contributions:
\begin{itemize}

\item We develop a data engine that labels 3D object parts from large-scale internet data to train a general-category model \textbf{without the need for human annotation}. Our data engine creatively combines existing vision and language foundation models.

\item We build the first model for 3D segmentation that is simultaneously \textbf{open-world}, \textbf{cross-category}, \textbf{part-level} and \textbf{feed-forward}. We achieve $\mathbf{260\%}$ \textbf{improvement in mIoU} and \textbf{6$\times$} to over \textbf{300$\times$} the inference speed compared to existing methods.
    
\item We release a \textbf{benchmark} for evaluating open-world 3D part segmentation for diverse objects, with \textbf{5$\times$} more unique part types than the largest existing benchmark. 
\end{itemize}

\begin{figure*}[h]
\begin{center}
\centerline{\includegraphics[width=2\columnwidth]{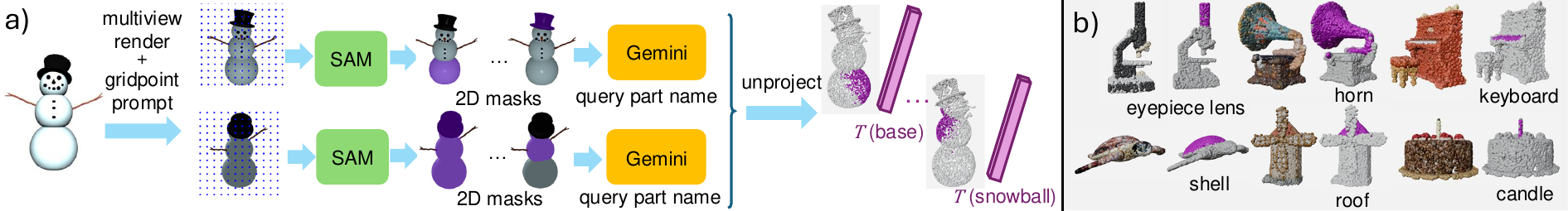}}
\caption{The \textbf{Data Engine}. (a) We render Objaverse assets into multiple views and pass each rendering to SAM with gridpoint prompts for segmentation. For each mask, we query Gemini for the corresponding part name, which gives us (mask, text) pairs. We embed the part name into the latent embedding space of a vision and language foundation model such as \siglip. We back-project mask pixels to obtain the points associated with each label embedding, yielding (points, text embedding) pairs. (b) Example annotations by the data engine.}
\label{fig:dataengine}
\end{center}
\vspace{-4mm}
\end{figure*}

\section{Related Work}
\mypar{Closed-world 3D segmentation.}
3D segmentation has been studied primarily in a closed world and with a coarse granularity that cannot go below whole objects. In specific settings such as indoor scenes or self-driving, state-of-the-art models are starting to achieve better generalization by training on multiple datasets, such as Mask3D~\cite{schult2023mask3d} and the PointTransformer series~\cite{zhao2021point, wu2022pt2, wu2024point}. However, these models are still domain-specific, and can only segment whole objects rather than parts.
Part-level segmentation is less studied. Early efforts started with the \shapenetpart dataset~\cite{yi2016scalable}  (16 object classes, $\leq 6$ parts per object). \partnete introduces articulated objects but is still limited to only 45 categories. Due to the limited number of categories and shared orientations (\eg, chairs all facing right), state-of-the-art part-level models~\cite{qian2022pointnext, loizou2023cross} cannot generalize well. 
Our work tackles both the challenges of generalization and granularity -- our model is part-level, and can segment any object part in an open-world setting.

\mypar{3D aggregation methods based on 2D renderings.}
With the progress of vision language models in 2D image understanding, some works directly assemble these models to obtain an ``aggregated" 3D understanding \emph{without} training a 3D model. An exemplary aggregation method uses multiview renderings of 3D scenes or objects, obtains their features in 2D based on models like CLIP~\cite{radford2021learning}, SAM~\cite{kirillov2023segment}, or GLIP~\cite{li2022grounded}, and combines them in 3D based on projection geometry. On the whole object level, such methods include OpenMask3D~\cite{takmaz2023openmask3d}. On the part level, such methods include PointCLIP~\cite{zhang2022pointclip}, PointCLIPV2~\cite{zhu2023pointclip}, PartSLIP~\cite{liu2023partslip} and PartSLIP++~\cite{zhou2023partslip++} for point clouds, and SATR~\cite{abdelreheem2023satr} for meshes. These models lack 3D geometry information and suffer from inconsistency across views. Furthermore, these methods are slow because they perform many inferences and the aggregation logic at test time.
Our method, which predicts in 3D with a single inference, is significantly faster. Our method also achieves stronger performance and better robustness to pose changes by leveraging 3D geometry information.

\mypar{Test-time optimization.}
Test-time optimization methods combine features from 2D models with a 3D representation, such as NeRF or Gaussian Splatting. At test time, these methods optimize the 3D representation with the 2D-sourced features attached. LERF~\cite{kerr2023lerf}, Distilled Feature Field~\cite{shen2023f3rm}, and Garfield~\cite{kim2024garfield} are based on radiance fields. Feature3DGS~\cite{zhou2024feature} is based on Gaussian splatting. These methods need to be optimized \emph{per scene} (or \emph{per object}), which can be slow (several minutes). Moreoever, their part-level capabilities have not been well-studied. Our method, feed-forward in nature, provides much faster inference with better performance.

\mypar{Distillation methods.}
Distillation methods train 3D models using 2D annotations. Generalization is a key limitation in prior works -- distillation is usually performed per dataset, even per category. OpenScene~\cite{peng2023openscene}, a whole-object segmentation model for indoor scenes, is distilled per dataset. For part segmentation, PartDistill~\cite{umam2024partdistill} is distilled \emph{per category}. Such models cannot perform inference zero-shot on unseen object classes, which is critical in real-world use cases. Our approach can be considered a distillation method that tackles the challenge of zero-shot generalization.
\section{Method}
\label{sec:method}
\begin{figure*}[h]
\begin{center}
\centerline{\includegraphics[width=2\columnwidth]{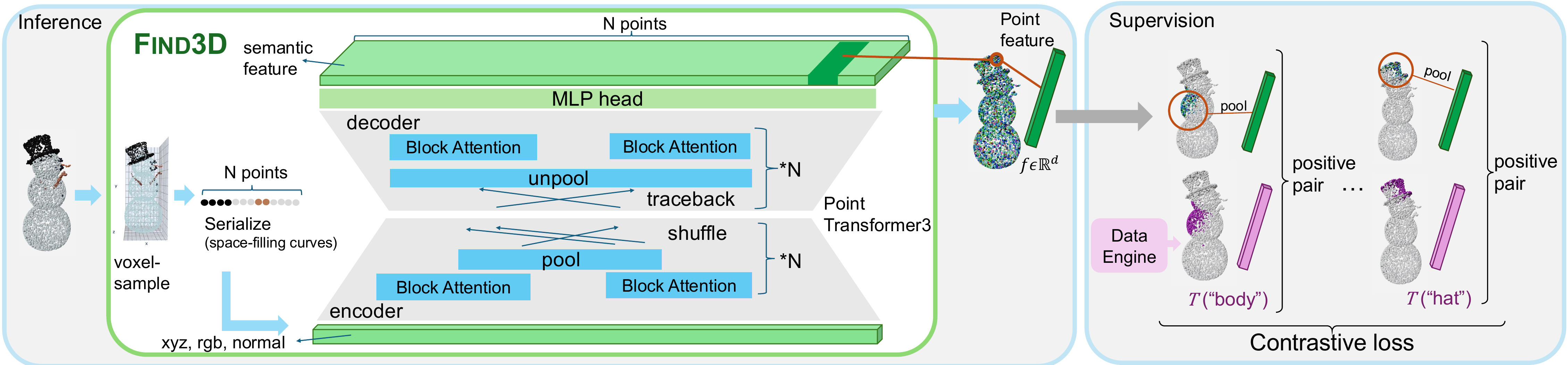}}
\caption{\method: an open-world part segmentation model. \method takes in a point cloud, voxelizes and serializes the points via space-filling curves into a sequence. The sequence is passed through a transformer architecture which returns a pointwise feature that is in the embedding space of a vision and language foundation model, denoted by $\mathbb{T}$. These features can be queried with any free-form text. \method is trained with a contrastive objective. For each (points, text embedding) label from the data engine, we use the averaged feature of these points as the predicted embedding, and pair it with the text embedding to form a positive pair in the contrastive loss.}
\label{fig:method}
\vspace{-4mm}
\end{center}
\end{figure*}
We propose a method, \method, to locate any object part in 3D based on a free-form language description, such as ``the wheel of a car".
As shown in \cref{fig:schematic} (panel b), we design a model that takes in a point cloud and outputs a queryable semantic feature for every point. This semantic feature is in the latent embedding space of a pre-trained CLIP-like~\cite{radford2021learning} model, such as \siglip~\cite{zhai2023sigmoid}. For any text query, we embed the query using the same model and calculate its cosine similarity with each point’s feature. This yields a pointwise similarity score that reflects the confidence of the part being located at that point. This score can be used to segment the object or localize specific parts.

Formally, given a point cloud $C = \{\mathbf{p_1}, ...{\mathbf{p_n}}\}$ with color and normals, for any point $\mathbf{p_i}=(x,y,z, n_x, n_y, n_z, r, g, b) \in \mathbb{R}^9$, we want to find a semantic feature $f_i \in \mathbb{R}^d$ which belongs in the same latent embedding space as a CLIP-like model, \eg, \siglip. At inference time, for any text $s$, we can get its \siglip embedding $T(s)$ and compute its cosine similarity with $f_i$, $cos(T(s), f_i)$.
For segmentation, \method assigns each point to the text query with the highest cosine similarity, and assigns ``no label" if all queries yield negative similarity scores.

\subsection{Data Engine}
Obtaining large-scale 3D annotations for generic object categories with human-in-the-loop pipelines is onerous. We develop a scalable data engine that leverages annotations from 2D foundation models and geometrically unprojects them to 3D.

As illustrated in ~\cref{fig:dataengine}, our data engine leverages SAM~\cite{kirillov2023segment} and Gemini~\cite{reid2024gemini} to annotate 3D assets from Objaverse~\cite{deitke2023objaverse}. Since Objaverse assets do not have a fixed orientation, and Gemini provides higher-quality labels to objects seen in familiar orientations, we first prompt Gemini to select the best orientation based on 10 renderings (from different camera angles) of an object in each orientation.
For the chosen orientation, we pass all renderings to SAM with grid point prompts.
We discard masks that are too small (less than 350 pixels out of a 500$\times$500 image), too large (greater than 20\% of all pixels), or with low confidence from SAM. We overlay each mask on the original image and ask Gemini to name the shaded part. Prompts are detailed in the appendix. Masks with the same label are merged. This process generates labeled (mask, text) pairs.
We map each mask to a set of points in the point cloud based on projection geometry. To make the point features queryable by language, we align point features to the language embedding space of a pretrained model, such as \siglip. We embed the label texts and use the text features as supervision.

The data engine processes 36044 Objaverse objects under LVIS categories selected by ~\cite{long2024wonder3d, wonder3drepo}. Each part can be annotated differently from different views, denoting various aspects of part, such as location (\eg, ``bottom"), material (\eg, ``snowball"), and function (\eg, ``body"). Labels also have different levels of granularity. For example, in \cref{fig:dataengine}, one granularity is individual snowballs, and another granularity is the whole snowman. The diversity of our labels helps the model handle the inherent ambiguity in segmentation. Our data engine annotates 30K objects from 761 unique categories with 2.1 million parts in total. Our annotations contain 124615 unique part types, which is over $\mathbf{1775\times}$ the number of unique part types in existing datasets combined (\shapenetpart and \partnete contain 71 unique part types in total). \cref{fig:dataengine} panel b shows some example annotations covering a wide range of part types and object geometries. We provide more annotation examples by our data engine in the appendix.

\subsection{Open-World 3D Part Model}
\label{sec:model}

\mypar{Architecture.}
\method adopts the PT3~\cite{wu2024point} architecture that treats point clouds as sequences, as illustrated in~\cref{fig:method}. To align the point features into the latent embedding space of \siglip, we append a lightweight 4-layer MLP to the last layer of the transformer. This returns a 768-dimension feature per point.
Our model contains 46.2 million parameters.

\mypar{Training.}
Leveraging the diverse annotations from our data engine requires some care. We cannot define a direct pointwise loss because: 1) The same point can have multiple labels that denote various aspects of a part such as location, material, and function. Some labels may also be incorrect; and 2) Many points are unlabeled - as shown in \cref{fig:method} (right), each mask only labels points visible from one camera view, and thus parts are likely to be labeled partially.

The challenge of partial labels can be resolved if the model can map features based on 3D geometry: points on the same ball should share similar features, and if we align the features of some points on that ball correctly to the text embedding, the other points' features should also be aligned. The challenge of multiple labels can be resolved by the contrastive formulation: each point's feature is encouraged to be close to the embeddings of all its labels, which allows for flexible text queries at inference time.
As illustrated in ~\cref{fig:method} (right), we define the contrastive pairing as follows: for each label, the ground truth is the \siglip embedding of the text. The predicted value is the average feature of all points that correspond to the label. 
This pooling can also be regarded as a way to ``denoise" the labels -- while an individual point might be affected by conflicting or incorrect labels, it is unlikely that all points are subjected to the same error.

Formally, our data engine provides (points, text embedding) labels, which we denote as $(C_i, T(\text{label}_i))$
 where $C_i$ is a subset of the point cloud that this label corresponds to, and $T(\text{label}_i)$ is the label embedding. We denote the pooled feature from the labeled points as $f(C_i)$, where $f$ is our model. We define the contrastive loss as follows:

\begin{equation}
    l_i=-\log \frac{\exp(f(C_{i}) \cdot T(\text{label}_i))}{\sum_{j=1}^ {\vert\mathcal{B}\vert} \exp(f(C_{i}) \cdot T(\text{label}_j))}
\end{equation}
where $\mathcal{B}$ denotes all labels of all objects in a batch.
For training, we use a batch size of 64 objects, corresponding to $\sim3000$ positive pairs per batch.

To achieve generalization, in addition to training on diverse data provided by the data engine, we also apply data augmentations, including random rotation (implemented as sequential random rotation along all three axes), scaling, flipping, jittering, chromatic auto contrast, chromatic translation, and chromatic jitter. These augmentations help avoid over-reliance on object poses and color, and nudges the model to take up 3D geometric cues. 
We perform a 90:10 train-validation split on the 27552 objects provided by the data engine, and train with the Adam optimizer~\cite{Kingma2014AdamAM} with a cosine annealing learning rate schedule, starting at 0.0003 and ending at 0.00005 over 80 epochs.

\section{A General Open-World 3D Part Benchmark}
\label{sec:benchmark}
\begin{figure}[t]
\begin{center}
\centerline{\includegraphics[width=\columnwidth]{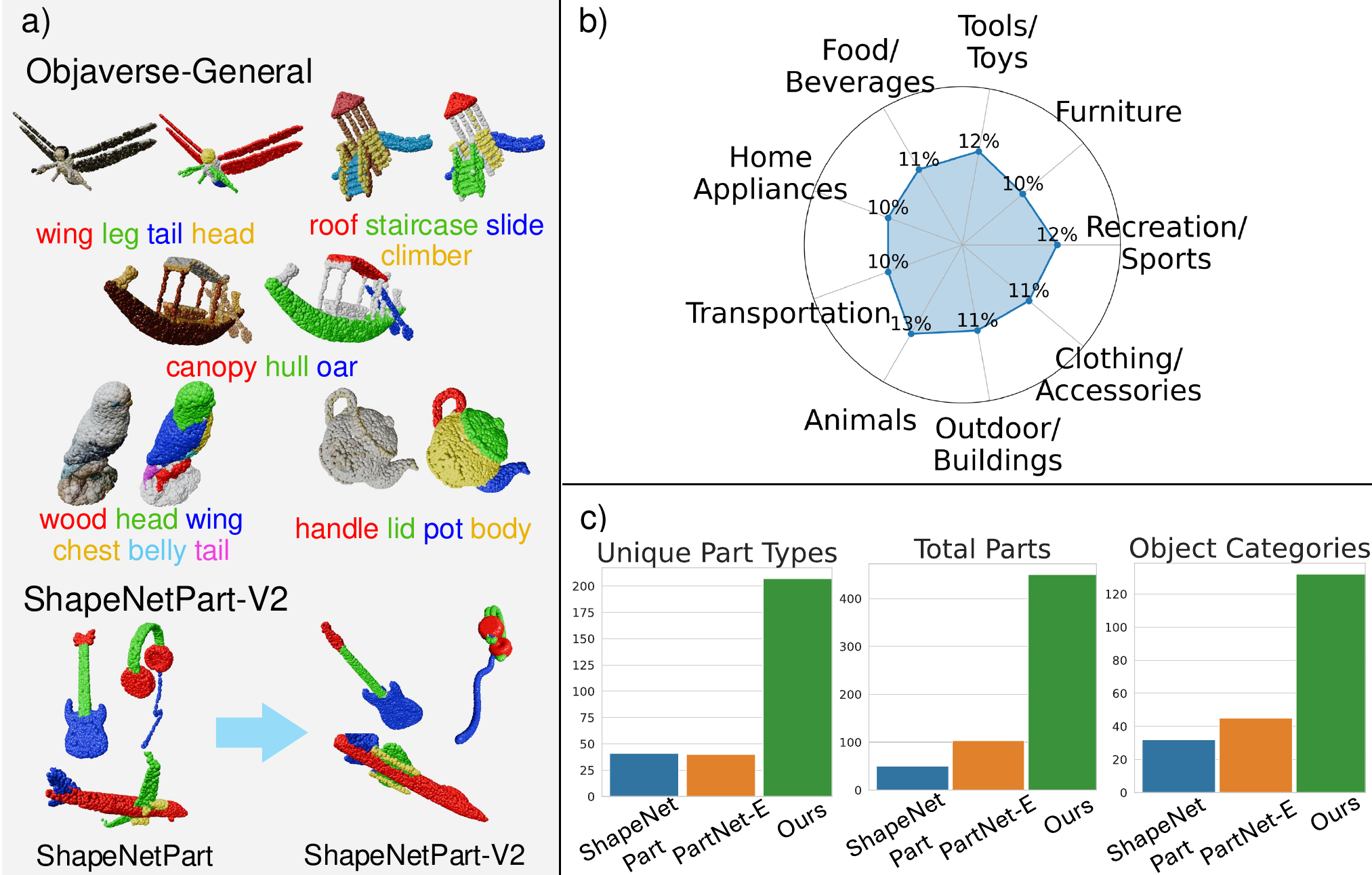}}
\caption{Our benchmark. (a) Examples of \objaversegeneral and \shapenetpartobj. \objaversegeneral contains diverse objects and parts, and \shapenetpartobj is sourced to look similar to \shapenetpart to test various methods' generalization capability. (b) Object category breakdown of \objaversegeneral, which covers 9 categories from tools to buildings. (c) Comparison with existing benchmarks. We have $\mathbf{5 \times}$ more unique part types, $\mathbf{4.4 \times}$ more total annotated parts, and $\mathbf{2.9\times}$ more object categories.}
\label{fig:benchmark}
\end{center}
\vspace{-9mm}
\end{figure}
Existing 3D part segmentation benchmarks only contain a small number of categories with a fixed set of parts, and are limited to narrow domains. \shapenetpart~\cite{yi2016scalable} contains 16 object categories with 41 unique part types, and the domain is limited to CAD models with canonical orientations (\eg, all chairs face right). \partnete~\cite{mo2019partnet, xiang2020sapien} contains 45 categories with 40 unique part types (\eg, ``button" is a common part across categories), and is restricted to simple home objects such as bottles and doors.
As shown in ~\cref{fig:benchmark}, we introduce a new human-annotated benchmark featuring a diverse range of objects, shapes, parts, and poses. We source our data from Objaverse~\cite{deitke2023objaverse}. Our benchmark contains 132 object categories, 450 total parts and 207 unique part types, over $\mathbf{5 \times}$ that of existing benchmarks, as shown in~\cref{fig:benchmark}. We hope this benchmark can advance 3D part segmentation towards more variable, “in-the-wild” scenarios.
The benchmark is divided into two sets:

\mypar{\objaversegeneral}
contains 100 objects with 350 parts from 100 diverse object categories, such as gondola, slide, lamppost, easel, penguin. 
These objects are in random orientations.
We hold out 50 out of the 100 categories from training, in order to evaluate out-of-distribution generalization to novel object types. The holdout categories are termed Unseen-Categories in Table \ref{table:miou_compare_all}.

\mypar{\shapenetpartobj} contains 32 objects from the same 16 object categories in \shapenetpart~\cite{yi2016scalable}. Inspired by ImageNetV2~\cite{recht2019imagenet}, we create this benchmark to evaluate generalization for models trained on \shapenetpart.

\section{Experiments}
\label{sec:experiments}

\begin{table}[t!]
\centering
\resizebox{0.99 \linewidth}{!}{
\begin{tabular}{l|c|c|c|c|c}
Method & Time & Open-world & Cross-category & Part-level & Feed-forward\\
\midrule
\method & \phantom{00}0.9s & \check & \check & \check & \check \\
PointCLIPV2 & \phantom{00}5.4s &\check &\check & \check & \cross \\
PartSLIP++ & 174.3s & \check & \cross$^\ast$ & \check & \cross \\
OpenMask3D & 296.5s &\check & \check & \cross & \cross \\
PartDistill$^\dagger$ & \phantom{00}0.7s (+348s) & \cross & \cross & \check & \check \\
PointNext  & \phantom{00}1.4s & \cross & \check & \check & \check \\
\bottomrule
\end{tabular}
} 
\caption{Properties and inference time of \method and baselines. \method is the only method that is open-world, cross-category, part-level and feed-forward. By ``feed-forward", we mean a method that performs direct 3D inference, without relying on a pipeline of multiview rendering, multiple 2D model inferences, and backprojection. Our method is $\mathbf{6\times}$ to $\mathbf{300\times}$ faster than other open-world models, and on par with closed-world models.\\$\ast$ PartSLIP++ finetunes a model per category. $\dagger$ PartDistill performs distillation for each new category (348s) and then does inference (0.7s). It is not open-world as the part names need to be defined prior to distillation. It only releases source code for two categories, and our reported speed is averaged across them.}
\label{table:method_properties}
\vspace{-3mm}
\end{table}

\begin{figure*}[t!]
\begin{center}
\centerline{\includegraphics[width=2\columnwidth]{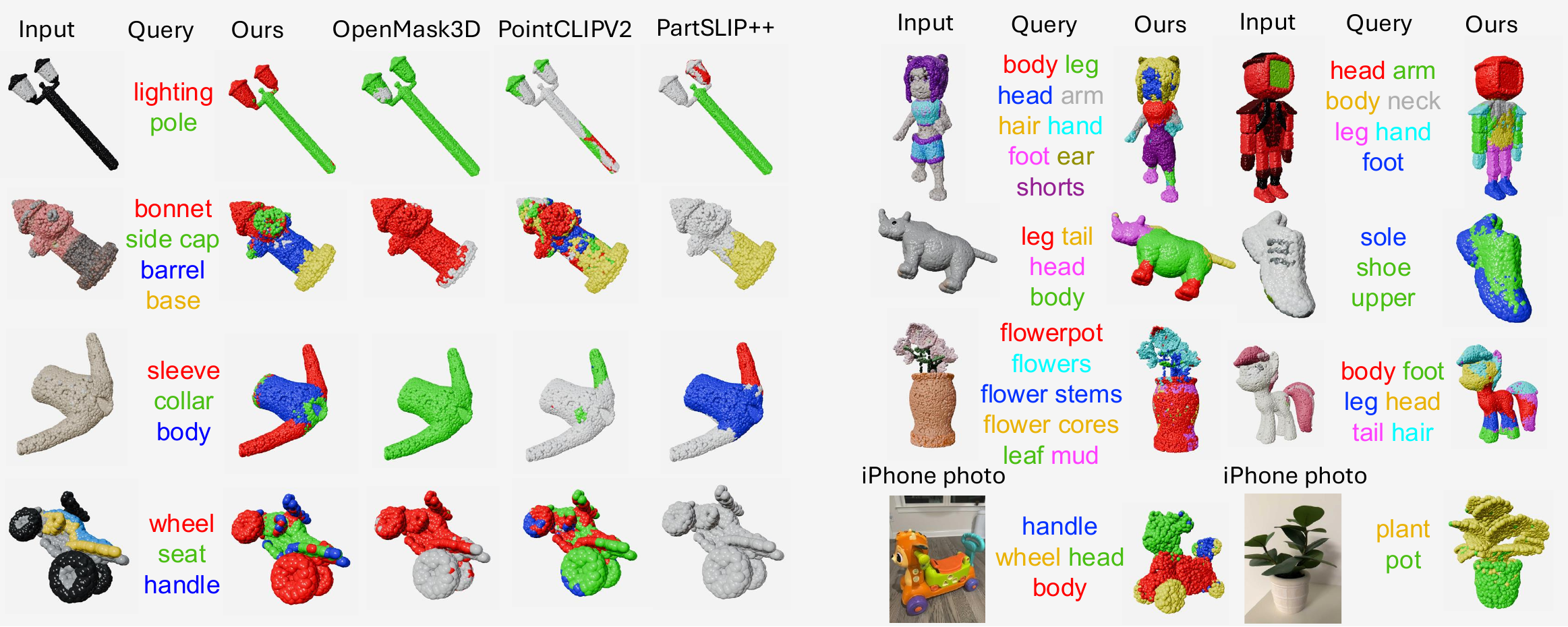}}
\caption{Qualitative results. Left: \method performs strongly on \objaversegeneral while baseline methods struggle. Right: more examples both from \objaversegeneral and PartObjaverse-Tiny, including out-of-distribution objects such as magical animals and complex anime-style characters. \method works on diverse object categories with up to 9 parts. It also generalizes to ``in-the-wild" iPhone photos (converted to point clouds via off-the-shelf image-to-3D method, as shown at bottom right.}
\label{fig:qualitative_comparison}
\end{center}
\vspace{-8mm}
\end{figure*}

As summarized in Table \ref{table:method_properties}, \method is the first method that is simultaneously, open-world, cross-category, part-level and feed-forward. \method not only shows strong zero-shot generalization, but also outperforms existing methods on their own domain. Our experiments show:

\begin{itemize}
\item \method achieves strong performance on diverse objects, with $\mathbf{260\%}$ \textbf{improvement in mIoU} from existing methods. \method exhibits strong out-of-distribution generalization, whereas baseline methods perform poorly on datasets they are not trained on, as shown qualitatively in \cref{fig:qualitative_comparison} and quantitatively in \cref{table:miou_compare_all}, \cref{table:miou_partnete}, \cref{table:miou_closed}.
\item \method is robust to variations such as query prompt rephrasing, object rotation, and domain shift, whereas baselines are sensitive to these changes. This is shown in \cref{fig:robustness}, \cref{table:miou_partnete}, \cref{table:miou_closed}.
\item \method is the most efficient open-world method with \textbf{6x} to \textbf{300x} speed improvement, as shown in \cref{table:method_properties}.
\end{itemize}

\subsection{Experimental Settings}
\label{sec:expsettings}
\textbf{Benchmarks.} In addition to our proposed benchmark (Sec.~\ref{sec:benchmark}), we also evaluate on two commonly used datasets for 3D part segmentation:
\shapenetpart~\cite{yi2016scalable} (16 object categories) and \partnete~\cite{mo2019partnet} (45 object categories).
For both datasets, we evaluate on their test set both in the canonical pose and in a randomly rotated (around all axes) pose, which correspond to the \emph{Canonical} and \emph{Rotated} columns in \cref{table:miou_compare_all}, \cref{table:miou_partnete}.

\begin{table*}[t!]
\begin{center}
\resizebox{\linewidth}{!}{
\begin{tabular}
{l|SS|SS|SS|SS|SS}
\toprule
mIoU (\%) & \multicolumn{4}{c|}{\objaversegeneral} & \multicolumn{6}{c}{\shapenetpart} \\
\midrule
 & \multicolumn{2}{c|}{Seen Categories} & \multicolumn{2}{c|}{Unseen Categories} & \multicolumn{2}{c|}{Canonical} & \multicolumn{2}{c|}{Rotated} & \multicolumn{2}{c}{\shapenetpartobj} \\
 \midrule
  & {\{part\} of a \{object\}}  & \{part\}  & {\{part\} of a \{object\}} &\raggedleft \{part\} & {\{part\} of a \{object\}} &\raggedleft \{part\}  & {\{part\} of a \{object\}} & \{part\} & {\{part\} of a \{object\}} & \{part\} \\
\midrule
  \method (ours) &  \cellcolor[rgb]{0.8,0.8,0.8}\textbf{33.78} & \cellcolor[rgb]{0.8,0.8,0.8}\textbf{34.10} & \textbf{26.21} & \textbf{27.41} & \textbf{28.39} & \textbf{24.09} & \textbf{29.64} & \textbf{23.71} & \textbf{42.15} & \textbf{30.02}\\
PointCLIPV2 & 9.81 & 11.27 & 10.27 & 11.09 & \cellcolor[rgb]{0.8,0.8,0.8}16.91 & \cellcolor[rgb]{0.8,0.8,0.8}20.22 & \cellcolor[rgb]{0.8,0.8,0.8}16.88 & \cellcolor[rgb]{0.8,0.8,0.8} 18.19 & 15.14 & 17.11 \\
PartSLIP++ & 2.69 & 15.03 & 0.57 & 10.43  & 1.43 & 6.46 & 0.94& 6.03 & 1.54 & 11.62 \\
OpenMask3D & 11.81 & 11.93 & 7.01 & 10.31 & 8.94 & 10.37 & 6.75 & 14.56 & 15.87 & 13.77\\
\bottomrule
\end{tabular}
} 
\end{center}
\vspace{-3mm}
\caption{Performance comparison with open-world methods on \objaversegeneral and \shapenetpart. \colorbox{gray!35}{Shaded} cells mean the method is trained on the same dataset (expected higher than white cells); white cells mean zero-shot evaluation. \method performs best on Objaverse-General, with $\mathbf{260\%}$ \textbf{improvement in mIoU} on unseen categories where all methods are evaluated zero-shot. On \shapenetpart, \method's zero-shot performance even exceeds PointCLIPV2 which is trained on the this dataset. We show results evaluated with 2 common query prompts: ``\{part\} of a \{object\}" and ``\{part\}" for all methods.}
\vspace{-3mm}
\label{table:miou_compare_all}
\end{table*}

\mypar{Metric.} 
We report class-average intersection-over-union (mIoU) as our metric, which is the mean IoU for all labeled parts per object, averaged across all object categories.

\mypar{Competing Methods}

\mypar{\underline{\textit{Open-world Baselines:}}}
\textbf{PointCLIPV2}~\cite{zhu2023pointclip} is an open-world 2D-to-3D pipeline involving multiple invocations of CLIP~\cite{radford2021learning}. It uses top-k prompts ($k =1400 \times n_\text{parts}$ per object) selected on the test set of \shapenetpart.
\textbf{PartSLIP++}~\cite{zhou2023partslip++} is a detection-based pipeline involving invocations of GLIP~\cite{li2022grounded} and a custom algorithm for finding superpoints. It finetunes a separate model for each category in \partnete. We evaluate its zero-shot checkpoint for fairness of comparison.
\textbf{OpenMask3D}~\cite{takmaz2023openmask3d} is an open-vocabulary, 2D-to-3D pipeline trained on scenes.

PointCLIPV2 and OpenMask3D are dense methods that assign a label to every point. We provide the text query ``other" as an option for no label on benchmarks that contain unlabeled points.

\mypar{\underline{\textit{Closed-world Baselines:}}}
\textbf{PointNeXt}~\cite{qian2022pointnext} is a state-of-the-art closed-world point cloud segmentation model trained on \shapenetpart. Due to its closed vocabulary, it cannot be evaluated on other datasets.
\textbf{PartDistill}~\cite{umam2024partdistill} is a category-specific 2D-to-3D distillation method, which is open-world prior to distillation but closed-world at inference time. It cannot be evaluated on unseen object categories due to the category-specific nature of distillation. The code and data for this method are not fully released (only two categories are released). Since we cannot reproduce the approach, we show numbers claimed in the paper.

Because PartSLIP++ and OpenMask3D are slow (up to 5 minutes per object), they are infeasible to evaluate on the full test sets (evaluating OpenMask3D on \partnete test set would take 628 hours).
For fair evaluations of all methods, we create smaller subsets of 160 objects (10 objects/category $\times$ 16 categories) for \shapenetpart and 225 objects (5 objects/category $\times$ 45 categories) for \partnete.
For methods that are efficient to evaluate, we additionally report performance on the full test sets in the appendix. 
We observe the same rankings and similar results on the subsets and full sets.

\subsection{Experimental Results}

\mypar{Results on \objaversegeneral and \shapenetpart.}
\cref{table:miou_compare_all} reports the mIoU of \method and open-world baselines on \objaversegeneral and \shapenetpart. \method shows the strongest performance, with $\mathbf{260\%}$ \textbf{improvement in mIoU} compared to the best baseline, PointCLIPV2, when both are evaluated zero-shot out of distribution (\objaversegeneral -- Unseen Categories).
Additionally, even when evaluated zero-shot, \method outperforms PointCLIPV2 on \shapenetpart, the dataset it is trained on. 

\mypar{Qualitative results.}
As seen in \cref{fig:qualitative_comparison}, \method consistently outputs reasonable segmentations, while other methods struggle. PartSLIP++ is trained on \partnete with sparse part annotations, and thus tends to output ``no label" overly often. OpenMask3D struggles with the part-level granularity, and it usually only picks one part, or at most two parts, to represent the whole object. We additionally show examples in \objaversegeneral and PartObjaverse-Tiny \cite{yang2024sampart3dsegment3dobjects} for our method on the right. \method not only works with diverse objects and parts, but also generalizes to real-world iPhone photos (converted to point clouds with Trellis~\cite{xiang2024structured}, an off-the-shelf image-to-3D model), despite only being trained on synthetic assets.

\mypar{Results on \partnete.}
\begin{table}[t]
\centering
\begin{center}
\resizebox{0.99\linewidth}{!}{
\begin{tabular}{l|c|c|c|c}
\toprule
   mIoU(\%) & \multicolumn{2}{c|}{Canonical Orientation} & \multicolumn{2}{c}{Rotated}\\
\midrule
&\raggedleft\{part\} of a \{object\} &\raggedleft \{part\} &\raggedleft \{part\} of a \{object\} &\raggedleft \{part\} \arraybackslash\\
\midrule
\method (ours) & \textbf{16.86} & 16.38 & \textbf{17.62} & 17.16 \\
PartSLIP++ & \cellcolor[rgb]{0.8,0.8,0.8}\phantom{0}5.12 & \cellcolor[rgb]{0.8,0.8,0.8}\textbf{32.71} & \cellcolor[rgb]{0.8,0.8,0.8}\phantom{0}3.87 & \cellcolor[rgb]{0.8,0.8,0.8}\textbf{23.03} \\
PointCLIPV2 & 11.28& \phantom{0}9.70 & 10.32 & 10.22 \\
OpenMask3D & 12.54 & 11.24 & 11.93 & 11.67 \\
\bottomrule
\end{tabular}
} 
\end{center}
\vspace{-2mm}
\caption{Comparison of open-world methods on \partnete. \colorbox{gray!35}{Shaded} cells mean the method is trained on the same dataset (expected higher than white cells); white cells mean zero-shot evaluation. We evaluate with 2 prompt formats: ``\texttt{\{part\} of a \{object\}}" and ``\texttt{\{part\}}". PartSLIP++ achieves good performance with the ``\texttt{\{part\}}" prompt, but its performance drops $\mathbf{84\%}$ when we vary the query prompt. This dataset is challenging for our method due to the sparsity of labels and the presence of small parts that are not geometrically or colorfully prominent (\eg, buttons on a surface with the same color). 
Nevertheless, our method is more robust to rotation and prompt variation, and clearly outperforms the other baselines that are also evaluated zero-shot.}
\label{table:miou_partnete}
\vspace{-2mm}
\end{table}
\cref{table:miou_partnete} compares open-world methods on \partnete. PartSLIP++ is trained on this dataset while all other methods, including \method, are evaluated zero-shot, so the results in this table favor PartSLIP++. We evaluate on two common prompts: ``{part} of a {object}" (such as ``leg of a chair"), and ``part name" (``leg"). We see that PartSLIP++'s performance decreases greatly when we vary the query prompt, up to a $83\%$ drop. Our method, evaluated zero-shot, is more robust and outperforms PartSLIP++ with the ``{part} of a {object}" prompt. It also outperforms other zero-shot baselines under all evaluation configurations.

\mypar{Efficiency.}
As shown in ~\cref{table:method_properties}, \method only takes 0.9 seconds for inference, which is \textbf{6$\times$} to \textbf{300$\times$} faster than open-world baselines and on par with closed-world models. Inference time is the average per-object inference time on the \partnete subset evaluated on an A100.

\mypar{Comparing with closed-world methods on \shapenetpart and \shapenetpartobj.}
\cref{table:miou_closed} compares \method zero-shot with closed-world methods that are trained on \shapenetpart, which greatly favors the closed-world methods. PointNeXt is the leading closed-world method for this dataset, and PartDistill trains one model for each object category of \shapenetpart.
We additionally evaluate the methods' generalization capability on our \shapenetpartobj benchmark, similar to ImageNetV2~\cite{recht2019imagenet}. We see a $\mathbf{64\%}$ \textbf{drop} of PointNeXt. Even though PointNeXt is still in-domain and \method is evaluated out-of-distribution, \method shows a $\mathbf{1.5 \times}$ advantage. PartDistill is not reproducible and thus cannot be evaluated on \shapenetpartobj.

\begin{table}[t!]
\begin{center}
\resizebox{0.99\linewidth}{!}{
\begin{tabular}{l|c|S|S}
\toprule
mIoU (\%) & Trained on & {\shapenetpart} & {\shapenetpartobj} \\
\midrule
\method & Our data engine & 28.39 & 42.15\\
PointNeXt & \shapenetpart & \cellcolor[rgb]{0.8,0.8,0.8} 80.44 & 28.70 \\
PartDistill$^\dagger$ & \shapenetpart & \cellcolor[rgb]{0.8,0.8,0.8} 63.9 & N/A\\
\bottomrule
\end{tabular}
} 
\end{center}
\vspace{-5mm}
\caption{Performance comparison with closed-world methods. \colorbox{gray!35}{Shaded} cells mean the method is trained on the same dataset (expected higher than white cells); white cells mean zero-shot evaluation. PointNeXt, a state-of-the-art closed-world model, is trained on \shapenetpart, but its performance drops significantly on \shapenetpartobj. Our approach, which is trained on a domain different from either \shapenetpart and \shapenetpartobj, demonstrates a stronger out-of-domain performance ($1.5 \times$ better on \shapenetpartobj, +13.2\%). $\dagger$ PartDistill trains a model per-category on \shapenetpart. It does not release training source-code or checkpoints (apart from two categories), thus cannot be evaluated on \shapenetpartobj.}
\label{table:miou_closed}
\vspace{-0mm}
\end{table}

\subsubsection{Scaling Analysis}
\begin{figure}[t!]
  \centering
  \includegraphics[width=0.8\linewidth]{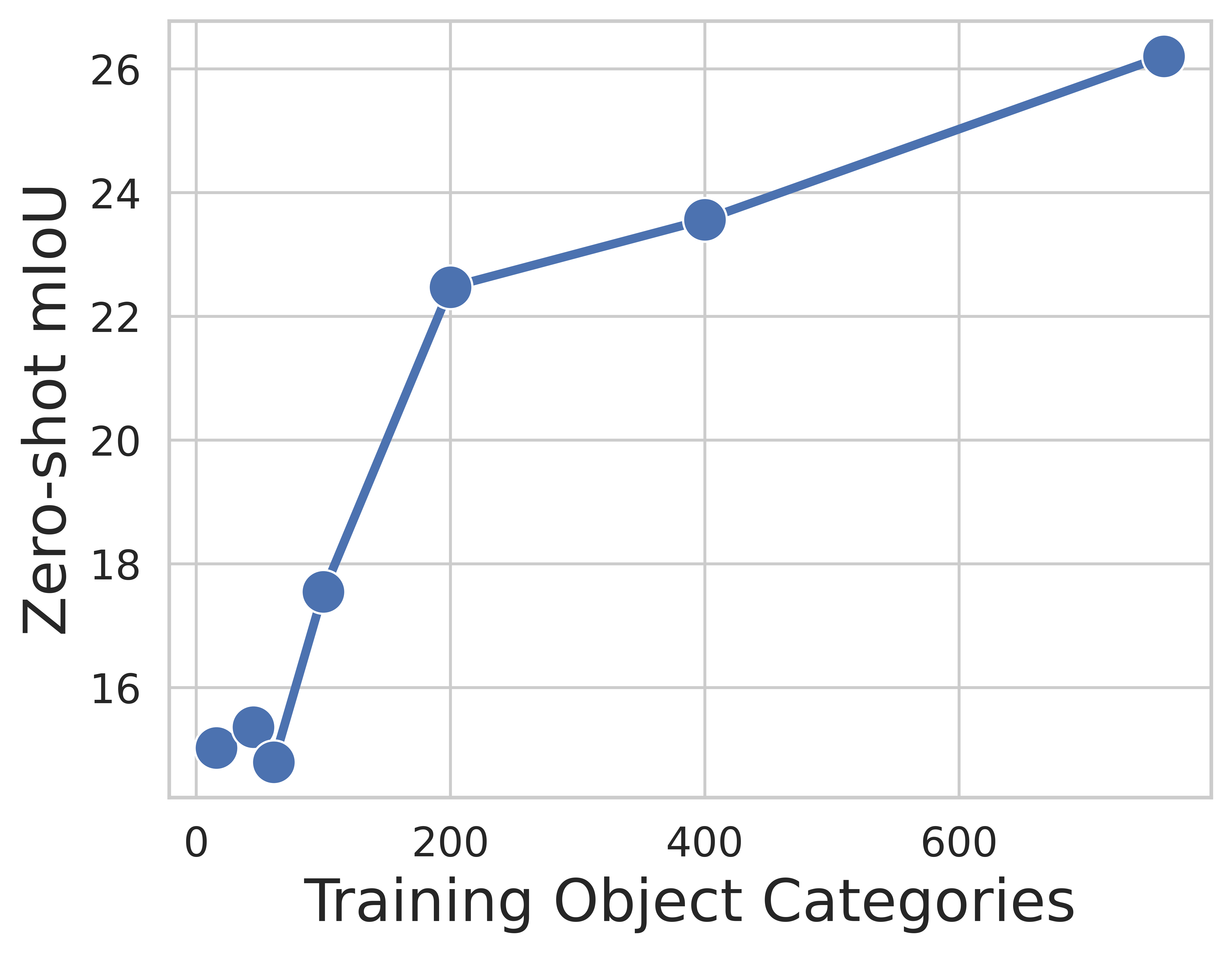}
  \vspace{-0.1in}
  \caption{Data scaling which shows that training on more object categories provides clear improvement in zero-shot mIoU. The evaluation is done on Objaverse-General Unseen Categories.}
   \label{fig:scaling}
   \vspace{-3mm}
\end{figure}
Data scaling is critical, as shown by the scaling analysis in~\cref{fig:scaling}. This finding highlights the importance of our data engine approach, which enables scaling in 3D.
We vary training object categories (x-axis) ranging from 16 categories (\shapenetpart dataset size), 45 categories (\partnete dataset size), all the way to 761 (our setting). We report zero-shot mIoU on \objaverseunseen (y-axis). We observe a strong scaling trend which is consistent with findings in many other data domains.

\subsubsection{Quantifying and Comparing Robustness}
\begin{figure}[h]
\begin{center}
\centerline{\includegraphics[width=\columnwidth]{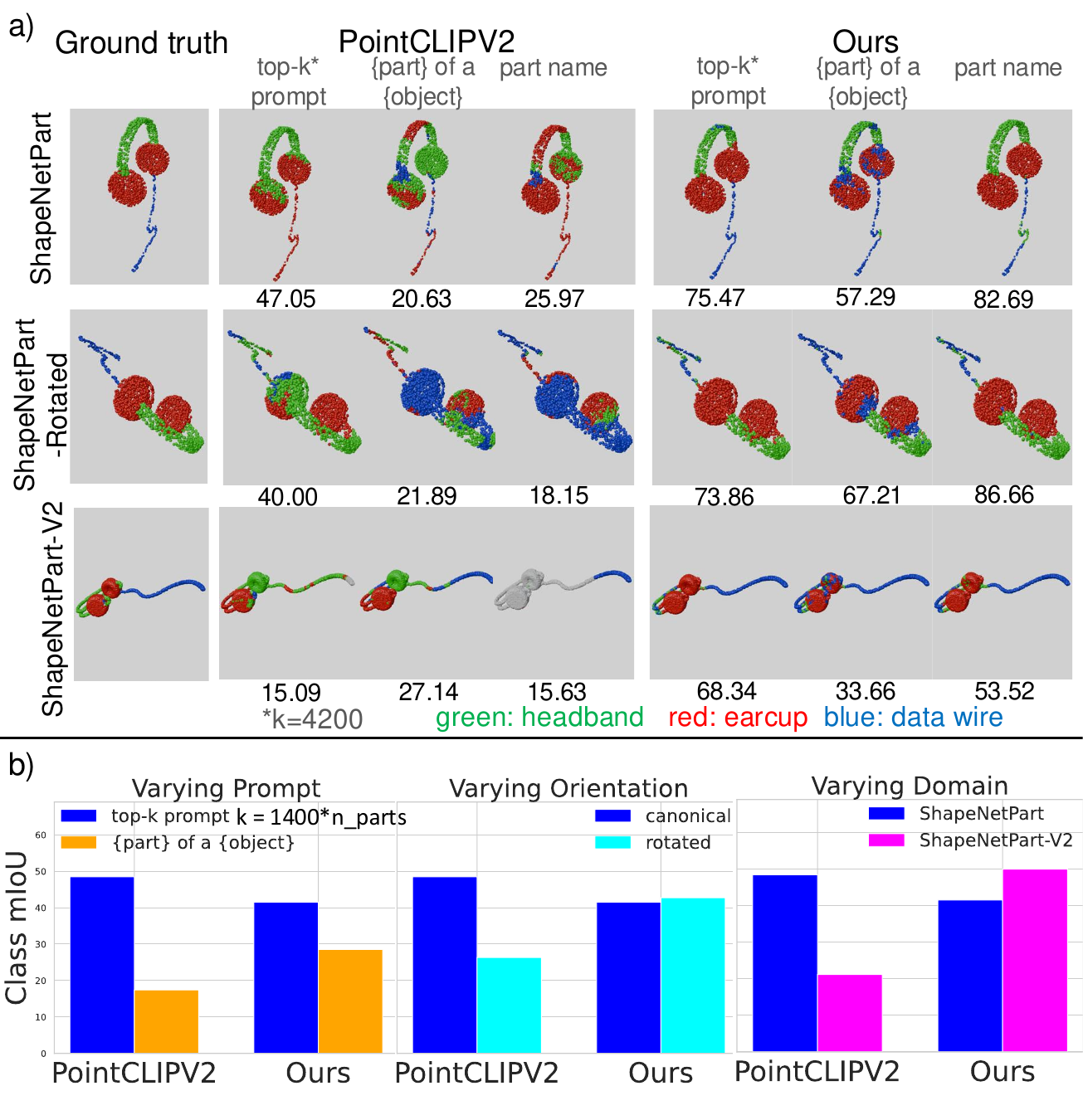}}
\vspace{-2mm}
\caption{(a) Qualitative comparison of PointCLIPV2 and \method on a \shapenetpart earphone (canonical and rotated) and a visually similar earphone from \shapenetpartobj. Top-k prompt reproduces evaluation in the PointCLIPV2 paper. PointCLIPV2's performance drops up to $68\%$, whereas our method stays consistent. (b) Comparison over all \shapenetpart categories. PointCLIPV2's performance drops $46\%$ to $64\%$ with varying conditions, while our method remains robust.}
\label{fig:robustness}
\end{center}
\vspace{-12mm}
\end{figure}

\cref{fig:robustness} evaluates the robustness of our method under different query text prompt, object orientation, and data domain, i.e. the data source of similarly-looking objects.

\mypar{Robustness to query prompt.}
PointCLIPV2 performs an extensive top-k prompt search on the test set: they iteratively optimize the prompt for each part (iteratively searching over 700 prompts per part and looping over all parts twice) and select the best prompt, i.e. iterate over $k=1400 \times n_{\text{part}}$ prompts and pick the one with the best test performance. For a fair comparison, we perform the same top-k search for our method. We also evaluate on two common prompts: ``{part} of a {object}" (such as ``leg of a chair"), and ``part name" (``leg"). As shown in \cref{fig:robustness}, with a change of prompt from top-k to ``{part} of a {object}", PointCLIPV2's performance drops from 48.47 to 17.42 ($64\%$ decrease), whereas our method exhibits more robust performance.

\mypar{Robustness to object orientation.}
We apply a random rotation by sampling three angles from $-\pi$ to $\pi$ and applying rotations along each of the X, Y, Z axis sequentially. PointCLIPV2's performance drops $46\%$ whereas our method does not drop but even increases $3\%$.

\mypar{Robustness to domain.}
We constructed \shapenetpartobj, a benchmark with objects from the same categories as \shapenetpart, but sourced from Objaverse assets. With this domain shift, PointCLIPV2's performance drops from 48.47 to 21.18 by $56\%$, whereas our method stays robust with a $20\%$ increase.

Comparison with other methods on other datasets in ~\cref{table:miou_compare_all} and ~\cref{table:miou_partnete} show similar trends.

\subsubsection{Flexibility of text queries}
\begin{figure}[t]
\begin{center}
\centerline{\includegraphics[width=\columnwidth]{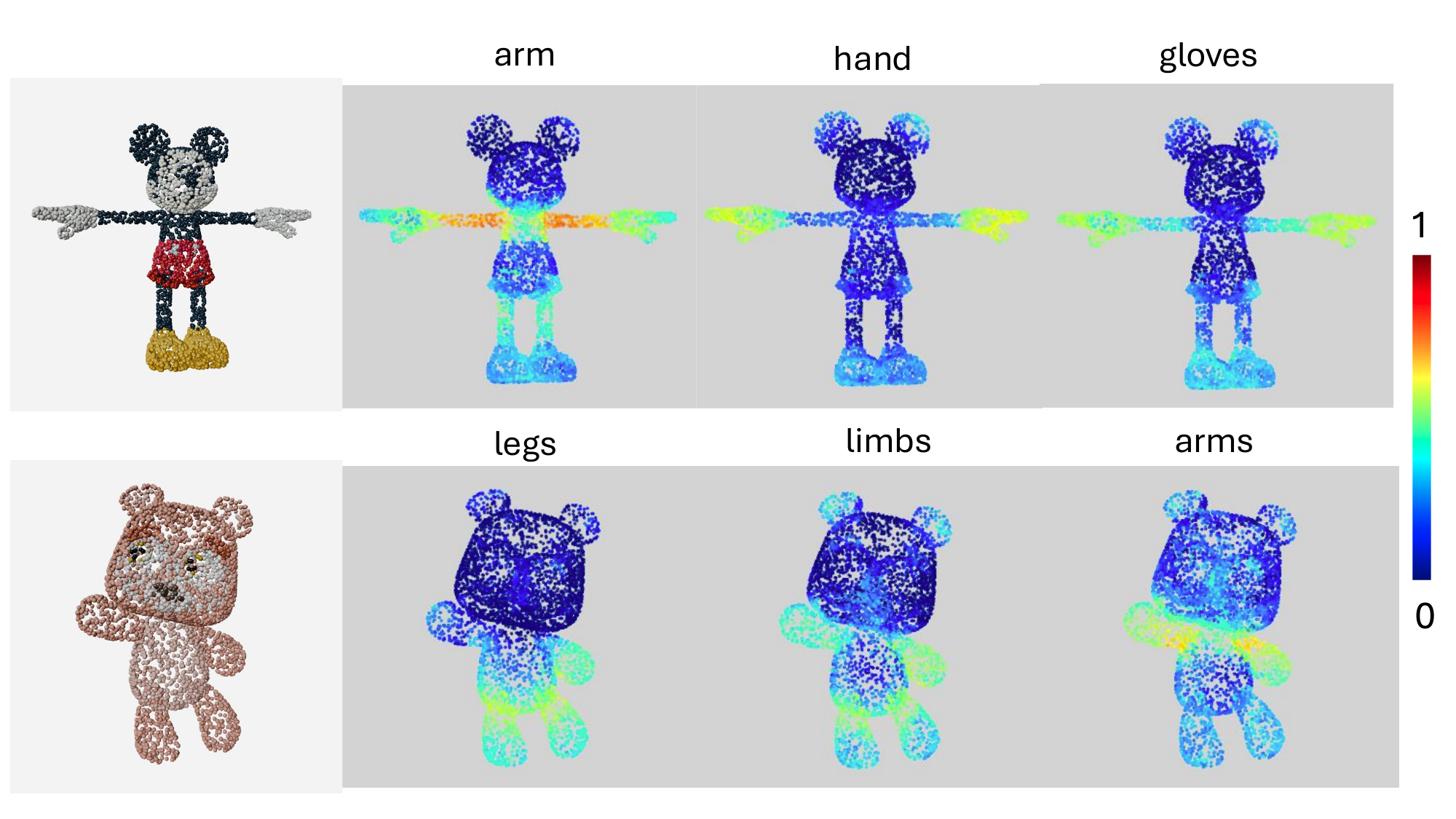}}
\vspace{-2mm}
\caption{Our method can support flexible text queries. For Mickey, one can either query by a body part such as ``hand" or by clothing such as ``gloves". For the teddy bear, one can either query the coarser-granularity concept ``limbs" or the finer-granularity ``arms" and ``legs".}
\label{fig:multi_granularity}
\end{center}
\vspace{-5mm}
\end{figure}

\method supports various query types that might occur in-the-wild. As shown in \cref{fig:multi_granularity}, \method can locate hands via different query types -- either by the body part ``hand" or by the clothing ``gloves". The teddy bear example demonstrates flexibility in query granularity -- one can query with ``limbs", a combination of arms and legs, or with ``arms" and ``legs" separately. For ease of visualization, the scores are min-clipped at 0.

\subsubsection{Failure Modes}

\begin{figure}[t]
\begin{center}
\centerline{\includegraphics[width=\columnwidth]{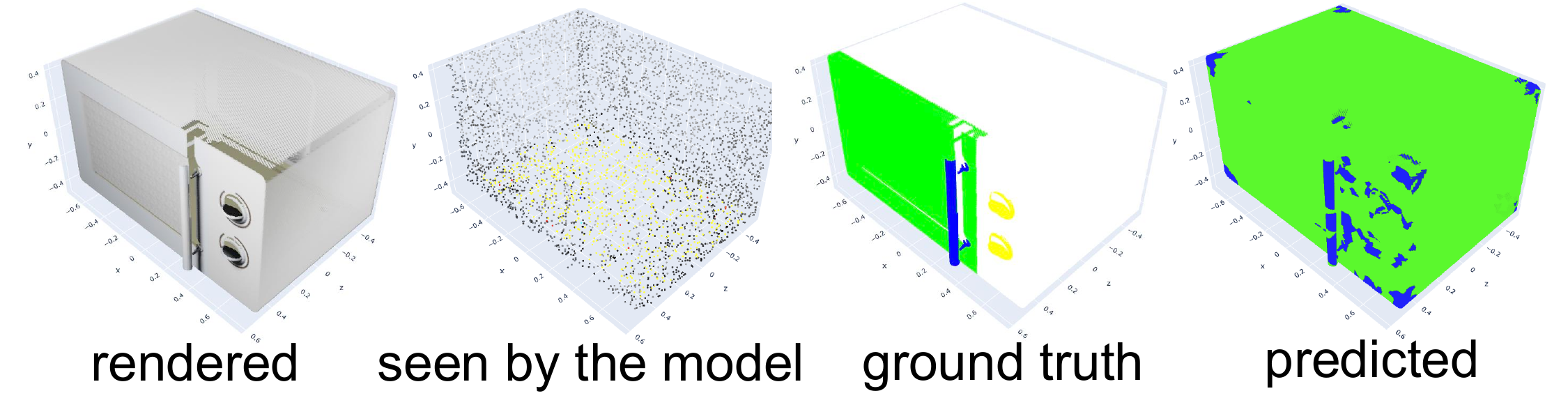}}
\vspace{-1mm}
\caption{A failure example. The leftmost image is a rendering of a microwave. The second image shows the point cloud at \method's sampled granularity, which loses most features.}
\label{fig:failure_mode}
\end{center}
\vspace{-10mm}
\end{figure}

We observe some limitations of \method: 1) Our model voxel-samples point clouds at the 0.02 resolution (after normalization). Fine-grained parts that are not geometrically prominent, such as bottons on a surface, are difficult for a point-cloud-only model like ours. 2) Because the model is trained to be rotational-equivariant, it tends to make symmetric predictions where all symmetric parts have the same label. \cref{fig:failure_mode} demonstrates an example from the \partnete dataset.
These limitations point to the complementary nature of the 2D and 3D modalities. While lacking in 3D geometry, the 2D modalities can better convey detailed appearance. Combining the image and the point cloud modality is a future direction.

\section{Discussions and Conclusions}
We present the first scaling study for 3D part segmentation. 
Key to our approach is a data engine that automatically annotates 3D assets from the internet, which allows us to train the first zero-shot generalist model for open-world 3D part segmentation on \emph{any} object.
Our method not only shows strong generalization, but even outperforms prior methods on the datasets they train on, despite being zero-shot.
We show that training object diversity is critical with a scaling analysis. We will release our code, benchmark and model checkpoints. We hope that by providing a diverse benchmark and the first demonstration of open-world 3D part segmentation at scale, we can encourage the community to shift away from customizations for small-scale datasets towards scale and generalization.
\section{Acknowledgments}
We would like to thank Ilona Demler, Raphi Kang, and Jiacheng Liu for feedback on the paper draft. We also thank Jiacheng Liu for help with the project demo. Ziqi Ma is supported by the Kortschak scholarship. This project is funded in part by NSF \#1918655, William H. Hurt Scholars Program, Powell Foundation, Google, and Amazon.
{
    \small
    \bibliographystyle{ieeenat_fullname}
    \bibliography{main}

\begin{thebibliography}{33}
\providecommand{\natexlab}[1]{#1}
\providecommand{\url}[1]{\texttt{#1}}
\expandafter\ifx\csname urlstyle\endcsname\relax
  \providecommand{\doi}[1]{doi: #1}\else
  \providecommand{\doi}{doi: \begingroup \urlstyle{rm}\Url}\fi

\bibitem[Abdelreheem et~al.(2023)Abdelreheem, Skorokhodov, Ovsjanikov, and Wonka]{abdelreheem2023satr}
Ahmed Abdelreheem, Ivan Skorokhodov, Maks Ovsjanikov, and Peter Wonka.
\newblock Satr: Zero-shot semantic segmentation of 3d shapes.
\newblock In \emph{Proceedings of the IEEE/CVF International Conference on Computer Vision}, pages 15166--15179, 2023.

\bibitem[Deitke et~al.(2023)Deitke, Schwenk, Salvador, Weihs, Michel, VanderBilt, Schmidt, Ehsani, Kembhavi, and Farhadi]{deitke2023objaverse}
Matt Deitke, Dustin Schwenk, Jordi Salvador, Luca Weihs, Oscar Michel, Eli VanderBilt, Ludwig Schmidt, Kiana Ehsani, Aniruddha Kembhavi, and Ali Farhadi.
\newblock Objaverse: A universe of annotated 3d objects.
\newblock In \emph{Proceedings of the IEEE/CVF Conference on Computer Vision and Pattern Recognition}, pages 13142--13153, 2023.

\bibitem[Kerr et~al.(2023)Kerr, Kim, Goldberg, Kanazawa, and Tancik]{kerr2023lerf}
Justin Kerr, Chung~Min Kim, Ken Goldberg, Angjoo Kanazawa, and Matthew Tancik.
\newblock Lerf: Language embedded radiance fields.
\newblock In \emph{Proceedings of the IEEE/CVF International Conference on Computer Vision}, pages 19729--19739, 2023.

\bibitem[Kim et~al.(2024)Kim, Wu, Kerr, Goldberg, Tancik, and Kanazawa]{kim2024garfield}
Chung~Min Kim, Mingxuan Wu, Justin Kerr, Ken Goldberg, Matthew Tancik, and Angjoo Kanazawa.
\newblock Garfield: Group anything with radiance fields.
\newblock In \emph{Proceedings of the IEEE/CVF Conference on Computer Vision and Pattern Recognition}, pages 21530--21539, 2024.

\bibitem[Kingma and Ba(2015)]{Kingma2014AdamAM}
Diederik~P. Kingma and Jimmy Ba.
\newblock Adam: {A} method for stochastic optimization.
\newblock In \emph{3rd International Conference on Learning Representations, {ICLR} 2015, San Diego, CA, USA, May 7-9, 2015, Conference Track Proceedings}, 2015.

\bibitem[Kirillov et~al.(2023)Kirillov, Mintun, Ravi, Mao, Rolland, Gustafson, Xiao, Whitehead, Berg, Lo, et~al.]{kirillov2023segment}
Alexander Kirillov, Eric Mintun, Nikhila Ravi, Hanzi Mao, Chloe Rolland, Laura Gustafson, Tete Xiao, Spencer Whitehead, Alexander~C Berg, Wan-Yen Lo, et~al.
\newblock Segment anything.
\newblock In \emph{Proceedings of the IEEE/CVF International Conference on Computer Vision}, pages 4015--4026, 2023.

\bibitem[Li et~al.(2022)Li, Zhang, Zhang, Yang, Li, Zhong, Wang, Yuan, Zhang, Hwang, et~al.]{li2022grounded}
Liunian~Harold Li, Pengchuan Zhang, Haotian Zhang, Jianwei Yang, Chunyuan Li, Yiwu Zhong, Lijuan Wang, Lu Yuan, Lei Zhang, Jenq-Neng Hwang, et~al.
\newblock Grounded language-image pre-training.
\newblock In \emph{Proceedings of the IEEE/CVF Conference on Computer Vision and Pattern Recognition}, pages 10965--10975, 2022.

\bibitem[Liu et~al.(2023)Liu, Zhu, Cai, Han, Ling, Porikli, and Su]{liu2023partslip}
Minghua Liu, Yinhao Zhu, Hong Cai, Shizhong Han, Zhan Ling, Fatih Porikli, and Hao Su.
\newblock Partslip: Low-shot part segmentation for 3d point clouds via pretrained image-language models.
\newblock In \emph{Proceedings of the IEEE/CVF conference on computer vision and pattern recognition}, pages 21736--21746, 2023.

\bibitem[Loizou et~al.(2023)Loizou, Garg, Petrov, Averkiou, and Kalogerakis]{loizou2023cross}
Marios Loizou, Siddhant Garg, Dmitry Petrov, Melinos Averkiou, and Evangelos Kalogerakis.
\newblock Cross-shape attention for part segmentation of 3d point clouds.
\newblock In \emph{Computer Graphics Forum}, page e14909. Wiley Online Library, 2023.

\bibitem[Long et~al.()Long, Guo, Lin, Liu, Dou, Liu, Ma, Zhang, Habermann, Theobalt, et~al.]{wonder3drepo}
Xiaoxiao Long, Yuan-Chen Guo, Cheng Lin, Yuan Liu, Zhiyang Dou, Lingjie Liu, Yuexin Ma, Song-Hai Zhang, Marc Habermann, Christian Theobalt, et~al.
\newblock Wonder3d objaverse subset uids.
\newblock \url{https://github.com/xxlong0/Wonder3D/blob/main/data_lists/lvis_uids_filter_by_vertex.json}.
\newblock Accessed: 2024-11-01.

\bibitem[Long et~al.(2024)Long, Guo, Lin, Liu, Dou, Liu, Ma, Zhang, Habermann, Theobalt, et~al.]{long2024wonder3d}
Xiaoxiao Long, Yuan-Chen Guo, Cheng Lin, Yuan Liu, Zhiyang Dou, Lingjie Liu, Yuexin Ma, Song-Hai Zhang, Marc Habermann, Christian Theobalt, et~al.
\newblock Wonder3d: Single image to 3d using cross-domain diffusion.
\newblock In \emph{Proceedings of the IEEE/CVF Conference on Computer Vision and Pattern Recognition}, pages 9970--9980, 2024.

\bibitem[Mo et~al.(2019)Mo, Zhu, Chang, Yi, Tripathi, Guibas, and Su]{mo2019partnet}
Kaichun Mo, Shilin Zhu, Angel~X Chang, Li Yi, Subarna Tripathi, Leonidas~J Guibas, and Hao Su.
\newblock Partnet: A large-scale benchmark for fine-grained and hierarchical part-level 3d object understanding.
\newblock In \emph{Proceedings of the IEEE/CVF conference on computer vision and pattern recognition}, pages 909--918, 2019.

\bibitem[Peng et~al.(2023)Peng, Genova, Jiang, Tagliasacchi, Pollefeys, Funkhouser, et~al.]{peng2023openscene}
Songyou Peng, Kyle Genova, Chiyu Jiang, Andrea Tagliasacchi, Marc Pollefeys, Thomas Funkhouser, et~al.
\newblock Openscene: 3d scene understanding with open vocabularies.
\newblock In \emph{Proceedings of the IEEE/CVF conference on computer vision and pattern recognition}, pages 815--824, 2023.

\bibitem[Qian et~al.(2024)Qian, Li, Peng, Mai, Al~Kader~Hammoud, Elhoseiny, and Ghanem]{qian2022pointnext}
Guocheng Qian, Yuchen Li, Houwen Peng, Jinjie Mai, Hasan~Abed Al~Kader~Hammoud, Mohamed Elhoseiny, and Bernard Ghanem.
\newblock Pointnext: revisiting pointnet++ with improved training and scaling strategies.
\newblock In \emph{Proceedings of the 36th International Conference on Neural Information Processing Systems}, 2024.

\bibitem[Radford et~al.(2021)Radford, Kim, Hallacy, Ramesh, Goh, Agarwal, Sastry, Askell, Mishkin, Clark, et~al.]{radford2021learning}
Alec Radford, Jong~Wook Kim, Chris Hallacy, Aditya Ramesh, Gabriel Goh, Sandhini Agarwal, Girish Sastry, Amanda Askell, Pamela Mishkin, Jack Clark, et~al.
\newblock Learning transferable visual models from natural language supervision.
\newblock In \emph{International conference on machine learning}, pages 8748--8763. PMLR, 2021.

\bibitem[Recht et~al.(2019)Recht, Roelofs, Schmidt, and Shankar]{recht2019imagenet}
Benjamin Recht, Rebecca Roelofs, Ludwig Schmidt, and Vaishaal Shankar.
\newblock Do imagenet classifiers generalize to imagenet?
\newblock In \emph{International conference on machine learning}, pages 5389--5400. PMLR, 2019.

\bibitem[Reid et~al.(2024)Reid, Savinov, Teplyashin, Lepikhin, Lillicrap, Alayrac, Soricut, Lazaridou, Firat, Schrittwieser, et~al.]{reid2024gemini}
Machel Reid, Nikolay Savinov, Denis Teplyashin, Dmitry Lepikhin, Timothy Lillicrap, Jean-baptiste Alayrac, Radu Soricut, Angeliki Lazaridou, Orhan Firat, Julian Schrittwieser, et~al.
\newblock Gemini 1.5: Unlocking multimodal understanding across millions of tokens of context.
\newblock \emph{arXiv preprint arXiv:2403.05530}, 2024.

\bibitem[Schult et~al.(2023)Schult, Engelmann, Hermans, Litany, Tang, and Leibe]{schult2023mask3d}
Jonas Schult, Francis Engelmann, Alexander Hermans, Or Litany, Siyu Tang, and Bastian Leibe.
\newblock Mask3d: Mask transformer for 3d semantic instance segmentation.
\newblock In \emph{2023 IEEE International Conference on Robotics and Automation (ICRA)}, pages 8216--8223. IEEE, 2023.

\bibitem[Shen et~al.(2023)Shen, Yang, Yu, Wong, Kaelbling, and Isola]{shen2023f3rm}
William Shen, Ge Yang, Alan Yu, Jansen Wong, Leslie~Pack Kaelbling, and Phillip Isola.
\newblock Distilled feature fields enable few-shot language-guided manipulation.
\newblock In \emph{7th Annual Conference on Robot Learning}, 2023.

\bibitem[Takmaz et~al.(2023)Takmaz, Fedele, Sumner, Pollefeys, Tombari, and Engelmann]{takmaz2023openmask3d}
Ay{\c{c}}a Takmaz, Elisabetta Fedele, Robert~W. Sumner, Marc Pollefeys, Federico Tombari, and Francis Engelmann.
\newblock {OpenMask3D: Open-Vocabulary 3D Instance Segmentation}.
\newblock In \emph{Advances in Neural Information Processing Systems (NeurIPS)}, 2023.

\bibitem[Umam et~al.(2024)Umam, Yang, Chen, Chuang, and Lin]{umam2024partdistill}
Ardian Umam, Cheng-Kun Yang, Min-Hung Chen, Jen-Hui Chuang, and Yen-Yu Lin.
\newblock Partdistill: 3d shape part segmentation by vision-language model distillation.
\newblock In \emph{Proceedings of the IEEE/CVF Conference on Computer Vision and Pattern Recognition}, pages 3470--3479, 2024.

\bibitem[Wu et~al.(2024{\natexlab{a}})Wu, Jiang, Wang, Liu, Liu, Qiao, Ouyang, He, and Zhao]{wu2024point}
Xiaoyang Wu, Li Jiang, Peng-Shuai Wang, Zhijian Liu, Xihui Liu, Yu Qiao, Wanli Ouyang, Tong He, and Hengshuang Zhao.
\newblock Point transformer v3: Simpler faster stronger.
\newblock In \emph{Proceedings of the IEEE/CVF Conference on Computer Vision and Pattern Recognition}, pages 4840--4851, 2024{\natexlab{a}}.

\bibitem[Wu et~al.(2024{\natexlab{b}})Wu, Lao, Jiang, Liu, and Zhao]{wu2022pt2}
Xiaoyang Wu, Yixing Lao, Li Jiang, Xihui Liu, and Hengshuang Zhao.
\newblock Point transformer v2: grouped vector attention and partition-based pooling.
\newblock In \emph{Proceedings of the 36th International Conference on Neural Information Processing Systems}, Red Hook, NY, USA, 2024{\natexlab{b}}.

\bibitem[Xiang et~al.(2020)Xiang, Qin, Mo, Xia, Zhu, Liu, Liu, Jiang, Yuan, Wang, Yi, Chang, Guibas, and Su]{xiang2020sapien}
Fanbo Xiang, Yuzhe Qin, Kaichun Mo, Yikuan Xia, Hao Zhu, Fangchen Liu, Minghua Liu, Hanxiao Jiang, Yifu Yuan, He Wang, Li Yi, Angel~X. Chang, Leonidas~J. Guibas, and Hao Su.
\newblock {SAPIEN}: A simulated part-based interactive environment.
\newblock In \emph{The IEEE Conference on Computer Vision and Pattern Recognition (CVPR)}, 2020.

\bibitem[Xiang et~al.(2024)Xiang, Lv, Xu, Deng, Wang, Zhang, Chen, Tong, and Yang]{xiang2024structured}
Jianfeng Xiang, Zelong Lv, Sicheng Xu, Yu Deng, Ruicheng Wang, Bowen Zhang, Dong Chen, Xin Tong, and Jiaolong Yang.
\newblock Structured 3d latents for scalable and versatile 3d generation.
\newblock \emph{arXiv preprint arXiv:2412.01506}, 2024.

\bibitem[Yang et~al.(2024)Yang, Huang, Guo, Lu, Wu, Lam, Cao, and Liu]{yang2024sampart3dsegment3dobjects}
Yunhan Yang, Yukun Huang, Yuan-Chen Guo, Liangjun Lu, Xiaoyang Wu, Edmund~Y. Lam, Yan-Pei Cao, and Xihui Liu.
\newblock Sampart3d: Segment any part in 3d objects, 2024.

\bibitem[Yi et~al.(2016)Yi, Kim, Ceylan, Shen, Yan, Su, Lu, Huang, Sheffer, and Guibas]{yi2016scalable}
Li Yi, Vladimir~G Kim, Duygu Ceylan, I-Chao Shen, Mengyan Yan, Hao Su, Cewu Lu, Qixing Huang, Alla Sheffer, and Leonidas Guibas.
\newblock A scalable active framework for region annotation in 3d shape collections.
\newblock \emph{ACM Transactions on Graphics (ToG)}, 35\penalty0 (6):\penalty0 1--12, 2016.

\bibitem[Zhai et~al.(2023)Zhai, Mustafa, Kolesnikov, and Beyer]{zhai2023sigmoid}
Xiaohua Zhai, Basil Mustafa, Alexander Kolesnikov, and Lucas Beyer.
\newblock Sigmoid loss for language image pre-training.
\newblock In \emph{Proceedings of the IEEE/CVF International Conference on Computer Vision}, pages 11975--11986, 2023.

\bibitem[Zhang et~al.(2022)Zhang, Guo, Zhang, Li, Miao, Cui, Qiao, Gao, and Li]{zhang2022pointclip}
Renrui Zhang, Ziyu Guo, Wei Zhang, Kunchang Li, Xupeng Miao, Bin Cui, Yu Qiao, Peng Gao, and Hongsheng Li.
\newblock Pointclip: Point cloud understanding by clip.
\newblock In \emph{Proceedings of the IEEE/CVF conference on computer vision and pattern recognition}, pages 8552--8562, 2022.

\bibitem[Zhao et~al.(2021)Zhao, Jiang, Jia, Torr, and Koltun]{zhao2021point}
Hengshuang Zhao, Li Jiang, Jiaya Jia, Philip~HS Torr, and Vladlen Koltun.
\newblock Point transformer.
\newblock In \emph{Proceedings of the IEEE/CVF international conference on computer vision}, pages 16259--16268, 2021.

\bibitem[Zhou et~al.(2024)Zhou, Chang, Jiang, Fan, Zhu, Xu, Chari, You, Wang, and Kadambi]{zhou2024feature}
Shijie Zhou, Haoran Chang, Sicheng Jiang, Zhiwen Fan, Zehao Zhu, Dejia Xu, Pradyumna Chari, Suya You, Zhangyang Wang, and Achuta Kadambi.
\newblock Feature 3dgs: Supercharging 3d gaussian splatting to enable distilled feature fields.
\newblock In \emph{Proceedings of the IEEE/CVF Conference on Computer Vision and Pattern Recognition}, pages 21676--21685, 2024.

\bibitem[Zhou et~al.(2023)Zhou, Gu, Li, Liu, Fang, and Su]{zhou2023partslip++}
Yuchen Zhou, Jiayuan Gu, Xuanlin Li, Minghua Liu, Yunhao Fang, and Hao Su.
\newblock Partslip++: Enhancing low-shot 3d part segmentation via multi-view instance segmentation and maximum likelihood estimation.
\newblock \emph{arXiv preprint arXiv:2312.03015}, 2023.

\bibitem[Zhu et~al.(2023)Zhu, Zhang, He, Guo, Zeng, Qin, Zhang, and Gao]{zhu2023pointclip}
Xiangyang Zhu, Renrui Zhang, Bowei He, Ziyu Guo, Ziyao Zeng, Zipeng Qin, Shanghang Zhang, and Peng Gao.
\newblock Pointclip v2: Prompting clip and gpt for powerful 3d open-world learning.
\newblock In \emph{Proceedings of the IEEE/CVF International Conference on Computer Vision}, pages 2639--2650, 2023.

\end{thebibliography}
}

\clearpage
\setcounter{page}{1}
\maketitlesupplementary
\renewcommand{\thesection}{A}

\section{Additional data engine annotation examples}
\begin{figure*}[ht!]
\begin{center}
\centerline{\includegraphics[width=1.5\columnwidth]{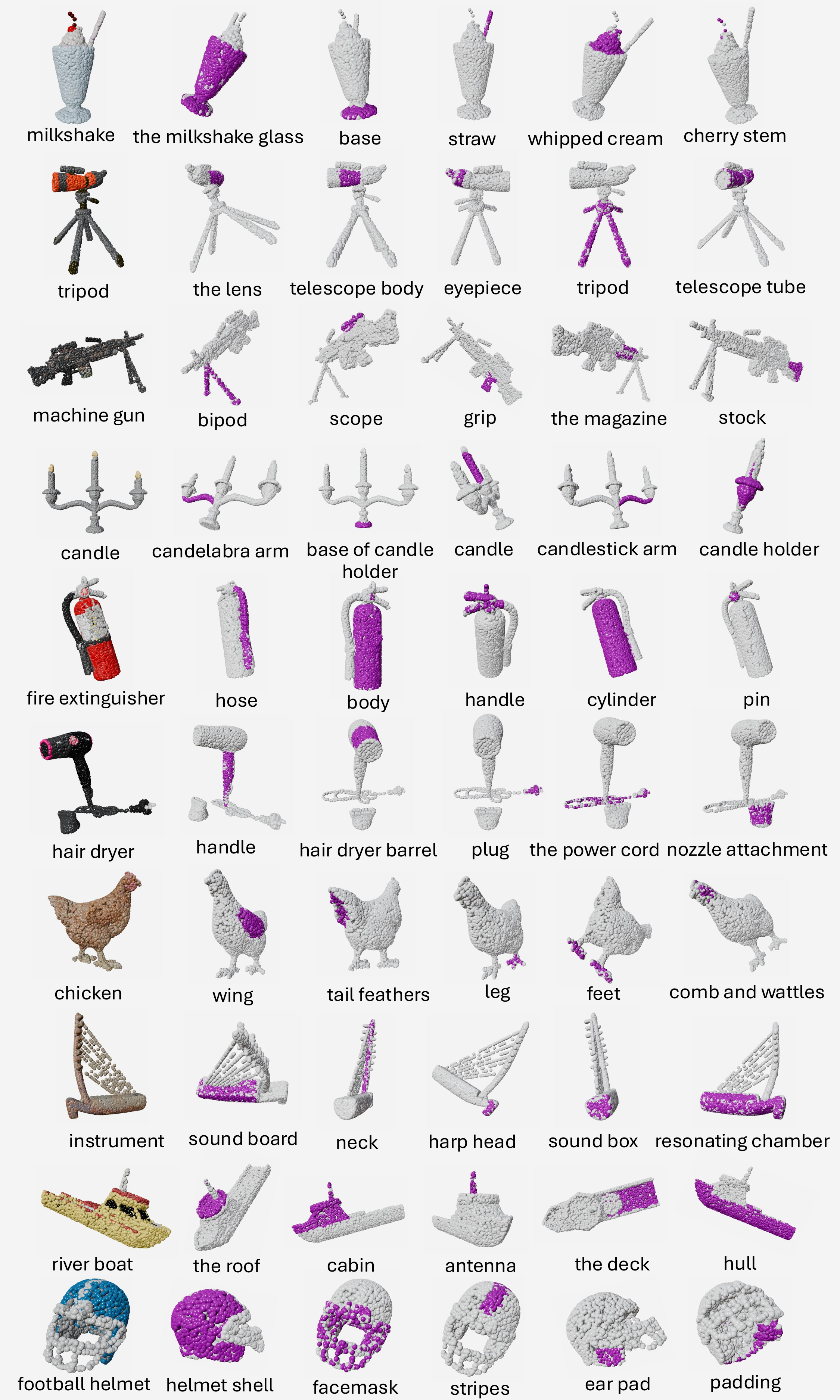}}
\caption{High-quality examples of data engine annotations. The LVIS label (from Objaverse) is shown below each input object. Our data engine annotates diverse objects and parts, including multiple captions for the same parts, such as ``candelabra arm" and ``candlestick arm", and multiple levels of granularity, such as ``helmet shell" and ``ear pad".}
\label{fig:engine_good}
\end{center}
\end{figure*}

\begin{figure*}[ht!]
\begin{center}
\centerline{\includegraphics[width=1.5\columnwidth]{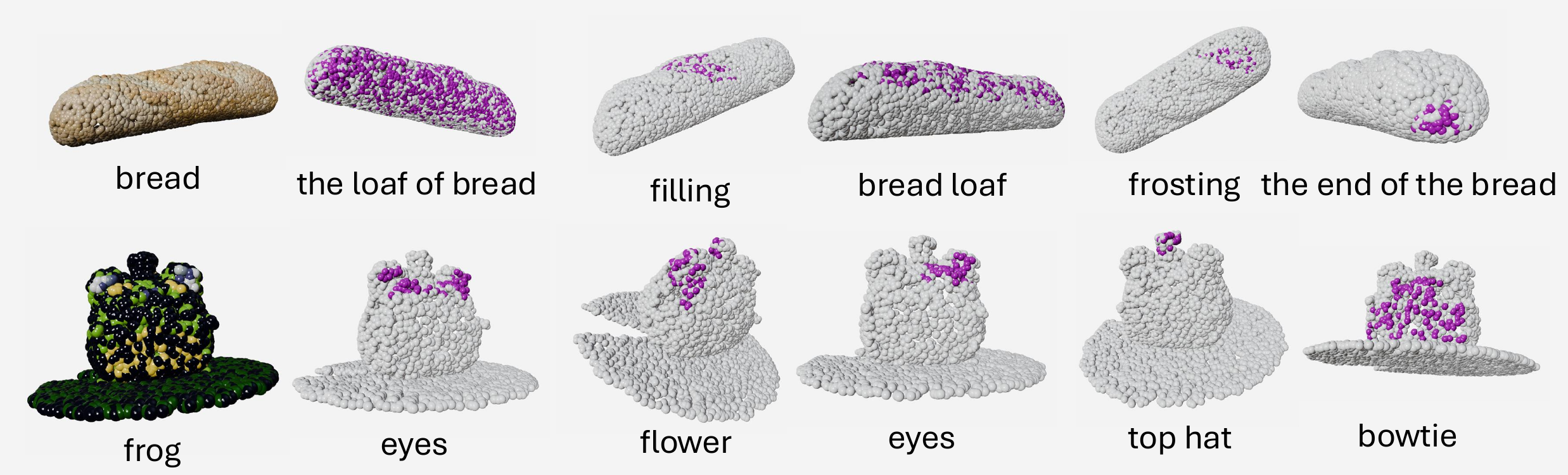}}
\caption{Lower-quality examples of data engine annotations. The LVIS label (from Objaverse) is shown below each input object. Some objects do not have pronounced parts, such as the baguette, and get partial part labels due to texture/lighting change on surfaces. Some objects are low quality, such as the cartoon frog, which results in incorrect labels.}
\label{fig:engine_bad}
\end{center}
\end{figure*}

We provide additional examples of our data engine annotations, both for high-quality examples in ~\cref{fig:engine_good} and lower-quality (but still useful) examples in ~\cref{fig:engine_bad}. Upon manual inspection of 50 randomly sampled objects, we observe $76\%$ high-quality examples. The annotations cover diverse objects and descriptions. For example, the body of a fire extinguisher is referred to both as ``body" and ``cylinder" from different views. The lower-quality examples are still useful for training -- they might not have pronounced parts, or contain partial masks (e.g., the baguette example), but the supervision signal still pushes the point features close to the correct semantic embedding (e.g. bread-related concepts). The low-quality cartoon frog contains both correct and incorrect masks. When learning from millions of such labels, the incorrect labels can be ``smoothed out" because it's unlikely that many frogs' bellies are all incorrectly labeled as ``bowtie".

\section{Additional qualitative examples of \method}
\begin{figure*}[ht]
\begin{center}
\centerline{\includegraphics[width=1.5\columnwidth]{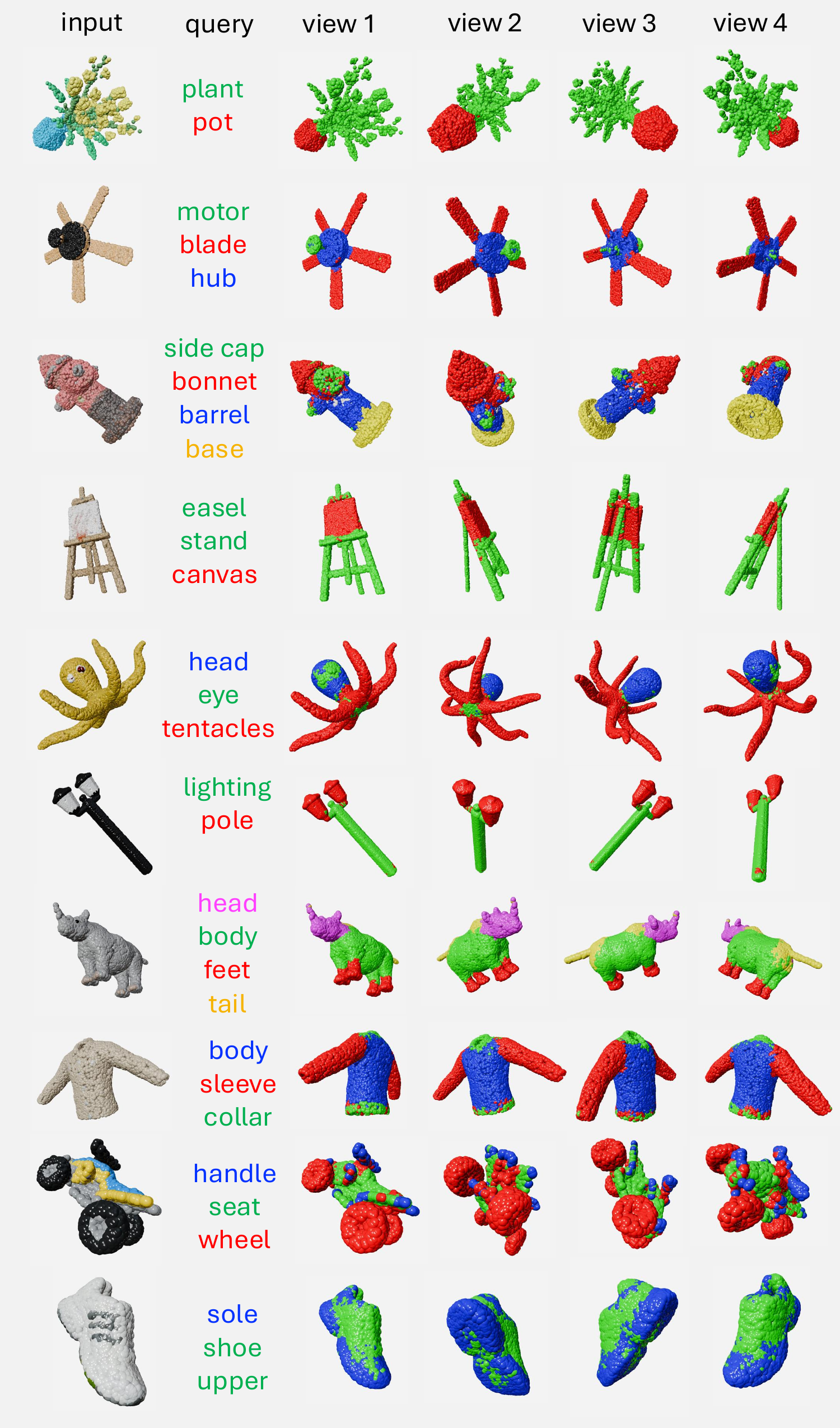}}
\caption{Multiple views of \method predictions on \objaversegeneral examples.}
\label{fig:qualitative_obj}
\end{center}
\end{figure*}

\begin{figure*}[ht]
\begin{center}
\centerline{\includegraphics[width=1.5\columnwidth]{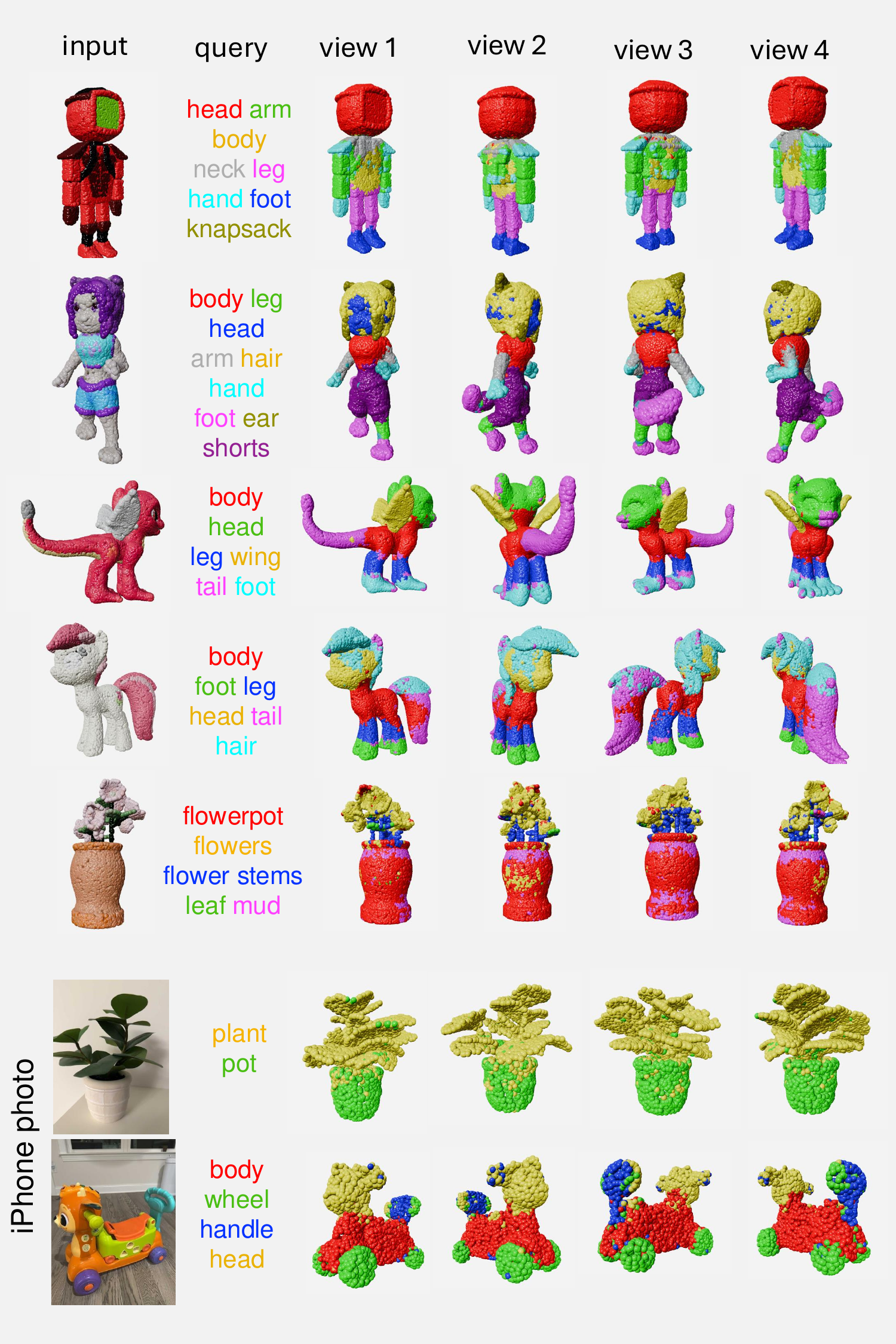}}
\caption{Multiple views of \method predictions on PartObjaverse-Tiny examples and iPhone photos (reconstructed to 3D with off-the-shelf method).}
\label{fig:qualitative_tinywild}
\end{center}
\end{figure*}

We provide additional qualitative results of \method in ~\cref{fig:qualitative_obj} and ~\cref{fig:qualitative_tinywild}. ~\cref{fig:qualitative_obj} shows predictions on \objaversegeneral from 4 views for each object.  ~\cref{fig:qualitative_tinywild} shows predictions on PartObjaverse-Tiny \cite{yang2024sampart3d} and iPhone photos (reconstructed to 3D via off-the-shelf single-image reconstruction method, Trellis \cite{xiang2024structured})). \method can segment diverse objects and parts, and can generalize to real-world objects, despite being trained on synthetic data.

\section{Experiments}
\label{sec:appendix_exp}
\subsection{Additional Results}
In \cref{table:miou_compare_all}, \cref{table:miou_partnete}, \cref{table:miou_closed} of the main paper, in order to evaluate all methods on the exact same data, we had to report results on subsets of \shapenetpart and \partnete because methods like PartSLIP++ and OpenMask3D are slow and infeasible to evaluate on the full test sets (\eg, OpenMask3D would take 628 hours on \partnete). Here we provide full-set results for methods that are feasible for full-set evaluation in \cref{table:miou_shapenetp_full} and \cref{table:miou_partnete_all}. The ranking of methods on the full sets and the subsets are the same.

\textbf{\shapenetpart.} \cref{table:miou_shapenetp_sub} compares all methods with various prompts, orientations, and data sources (\shapenetpart vs. \shapenetpartobj, a benchmark of the same object classes as \shapenetpart but sourced from Objaverse that we constructed, similar to ImageNetV2 \cite{recht2019imagenet}).
PointCLIPV2 is trained on this dataset, and other methods are evaluated zero-shot. \method performs the best in 8 out of 9 configurations, despite being zero-shot.
While \cref{table:miou_shapenetp_sub} reports metrics on the subset of \shapenetpart so that all methods can be evaluated strictly on the same dataset, for methods that are fast enough to evaluate on the full test set (\method and PointCLIPV2), we also report the full-set evaluation results in \cref{table:miou_shapenetp_full}. The full-set metrics are very close to the subset metrics. On the full set, we also see that \method performs better in 5 out of 6 settings.

On both the full set and the subset, \method, despite being zero-shot on this dataset, is the best-performing method in all configurations except for one---the canonical orientation with test-time top-k prompt searching. In this setting, PointCLIPV2, a method trained on this dataset and designed with test-time prompt searching in mind, performs slightly better. We note that this searching takes over an hour on an A100, which is unrealistic to perform in real applications. Our method is not designed for test-time prompt searching but clearly outperforms all baselines when doing direct inference.\\

\textbf{\partnete.} \cref{table:miou_partnete_all} shows results on \partnete, both on the subset (for all methods) and on the full set (for methods that are fast enough to evaluate on the full set). PartSLIP++, trained on this dataset, achieves the highest performance with the ``\texttt{\{part\}}" prompts, yet is very sensitive to prompt variation. We note that PartSLIP++ also releases category-specific checkpoints, but we use the cross-category checkpoint for fairness of comparison.
This dataset is more challenging for our method because many objects contain small parts that are not geometrically or colorfully prominent, such as buttons on a surface with the same color. Nevertheless, we see our method to be more robust to rotation and prompt variation, and clearly outperforms the other baselines that are not trained on this dataset. Furthermore, PartSLIP++ is a slow 2D-3D aggregation method, taking up to 3 minutes per object. Our method is over 30$\times$ faster.

\section{Data engine prompts}
\begin{figure}[t]
\begin{center}
\centerline{\includegraphics[width=\columnwidth]{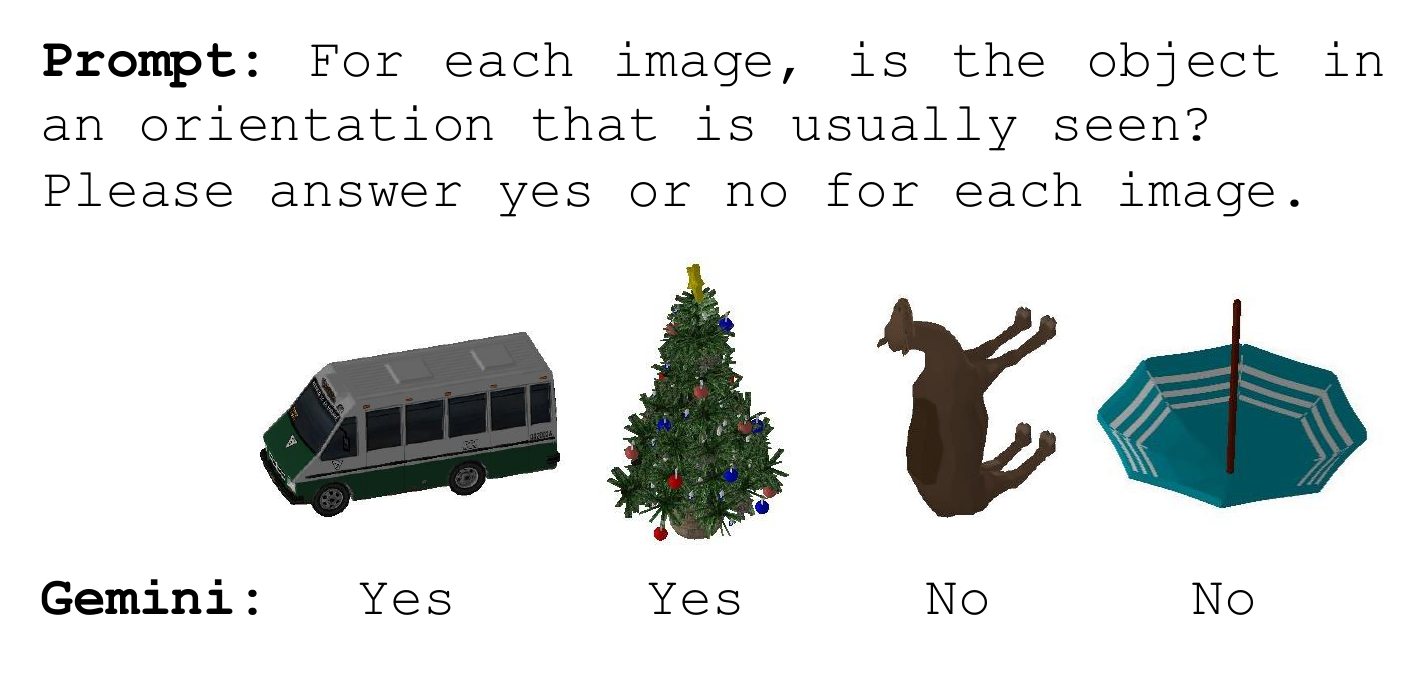}}
\caption{The prompt used to query Gemini for object orientation. The car and the Christmas tree are in common orientations (and thus will yield higher-quality annotations), whereas the camel and the parasol are not.}
\label{fig:gemini_orientation}
\vspace{-5mm}
\end{center}
\end{figure}

\begin{figure}[t!]
\begin{center}
\centerline{\includegraphics[width=\columnwidth]{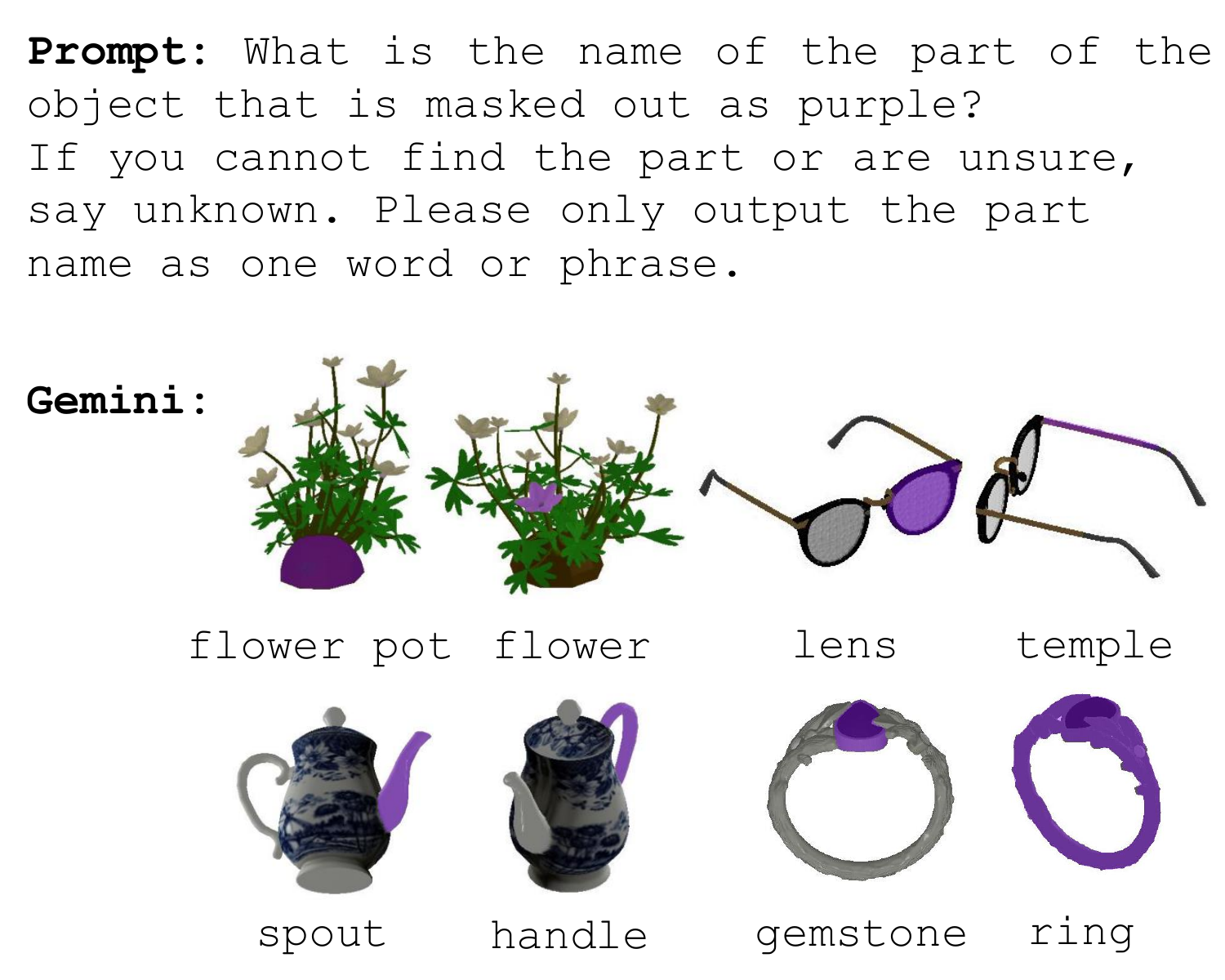}}
\caption{The prompt used to query Gemini for object part names. We show 2 example masks from different views for a potted plant, a pair of glasses, a teapot, and a ring.}
\label{fig:gemini_mask}
\vspace{-3mm}
\end{center}
\end{figure}

\cref{fig:gemini_orientation} shows the prompt we use to obtain object orientations from Gemini. For a given orientation, we render the object in 10 different views, and pass the prompt along with 10 renderings to Gemini. We calculate the percentage of ``yes" answers and choose the orientation with the highest ``yes" percentage. \cref{fig:gemini_orientation} also provides some example objects with answers from Gemini. \cref{fig:gemini_mask} shows the prompt we use to obtain part names from Gemini, along with some examples.

\FloatBarrier

\begin{table*}[h!]
\begin{center}
\begin{small}
\begin{tabular}{p{0.12\textwidth}|p{1.3cm}p{1.3cm}p{1.3cm}|p{1.3cm}p{1.3cm}p{1.3cm}|p{1.3cm}p{1.3cm}p{1.3cm}}
\toprule
   mIoU(\%) & \multicolumn{3}{c|}{\raggedleft Canonical Orientation} & \multicolumn{3}{c|}{\raggedleft Rotated} & \multicolumn{3}{c}{\raggedleft Objaverse-ShapeNetPart} \\
\midrule
& \raggedleft top-k & \{part\} of a \{object\} & \raggedleft \{part\} & \raggedleft top-k & \{part\} of a \{object\} & \raggedleft \{part\} & \raggedleft top-k & \{part\} of a \{object\} & \raggedleft \{part\}\arraybackslash\\
\midrule
PointCLIPV2 & \cellcolor[rgb]{0.8,0.8,0.8}\raggedleft \textbf{48.666} &\cellcolor[rgb]{0.8,0.8,0.8}\raggedleft 16.912 &\cellcolor[rgb]{0.8,0.8,0.8}\raggedleft 20.215 &\cellcolor[rgb]{0.8,0.8,0.8}\raggedleft 26.111 &\cellcolor[rgb]{0.8,0.8,0.8}\raggedleft 16.878 &\cellcolor[rgb]{0.8,0.8,0.8}\raggedleft 18.193 &\raggedleft 21.177 &\raggedleft 15.136 &\raggedleft 17.110 \arraybackslash\\
PartSLIP++ & \centering -- & \raggedleft 1.432 &\raggedleft 6.460 & \centering --& \raggedleft 0.937& \raggedleft 6.034& \centering -- & \raggedleft 1.542 & \raggedleft 11.622 \arraybackslash\\
OpenMask3D & \centering -- & \raggedleft 8.938& \raggedleft 10.373& \centering -- &\raggedleft 6.748 & \raggedleft 14.556  & \centering --& \raggedleft 15.870 & \raggedleft 13.768 \arraybackslash\\
\method (Ours) & \raggedleft 43.613& \raggedleft \textbf{28.386}& \raggedleft \textbf{24.085} & \raggedleft \textbf{43.781}&\raggedleft \textbf{29.637}& \raggedleft \textbf{23.712}&\raggedleft \textbf{50.002}& \raggedleft \textbf{42.151}& \raggedleft \textbf{30.018}\arraybackslash\\
\bottomrule
\end{tabular}
\end{small}
\end{center}
\caption{Detailed results on ShapeNet-Part subset. \colorbox{gray!35}{Shaded} cells mean the method is trained on the same dataset (expected higher than white cells), and white cells mean zero-shot evaluation. We evaluate different orientations, query prompts, and data domains (\shapenetpart vs. \shapenetpartobj). We evaluate on 3 types of prompts: ``\texttt{\{part\} of a \{object\}}", ``\texttt{\{part\}}", and top-k. Top-k prompt reproduces the PointCLIPV2 paper, which runs an iterative search over $1400 \times n_\text{parts}$ prompts per object category to choose the best query text prompts. For fairness of comparison, we follow the same procedure to get top-k prompt metrics, although our method is not designed with prompt searching in mind, and it is not realistic to conduct this slow ($>1$ hour on A100) searching process at inference time. Our method, despite being zero-shot on this dataset, has the best performance in 8 out of 9 configurations---all configurations except for the canonical orientation with top-k prompt searching.}
\label{table:miou_shapenetp_sub}
\end{table*}

\begin{table*}[h!]
\begin{center}
\begin{small}
\begin{tabular}{p{0.12\textwidth}|p{1.3cm}p{1.3cm}p{1.3cm}|p{1.3cm}p{1.3cm}p{1.3cm}}
\toprule
   mIoU(\%) & \multicolumn{3}{c|}{\raggedleft Canonical Orientation} & \multicolumn{3}{c}{\raggedleft Rotated} \\
\midrule
& \raggedleft top-k & \{part\} of a \{object\} & \raggedleft \{part\} & \raggedleft top-k & \{part\} of a \{object\} & \raggedleft \{part\} \arraybackslash\\
\midrule
PointCLIPV2 & \cellcolor[rgb]{0.8,0.8,0.8}\raggedleft \textbf{48.472} &\cellcolor[rgb]{0.8,0.8,0.8}\raggedleft 17.471 &\cellcolor[rgb]{0.8,0.8,0.8} \raggedleft 20.157  &\cellcolor[rgb]{0.8,0.8,0.8}\raggedleft 26.337 & \cellcolor[rgb]{0.8,0.8,0.8}\raggedleft 17.034 & \cellcolor[rgb]{0.8,0.8,0.8}\raggedleft 18.021\arraybackslash\\
\method (Ours) & \raggedleft 41.517& \raggedleft \textbf{28.532} & \raggedleft \textbf{23.569} & \raggedleft \textbf{42.734}& \raggedleft \textbf{29.966} & \raggedleft \textbf{23.794} \arraybackslash\\
\bottomrule
\end{tabular}
\end{small}
\end{center}
\caption{Detailed results on ShapeNet-Part full test set. \colorbox{gray!35}{Shaded} cells mean the method is trained on the same dataset (expected higher than white cells), and white cells mean zero-shot evaluation. PartSLIP2 and OpenMask3D are too slow and thus infeasible to evaluate on the full test set. The metrics are very close to the subset results in the previous table. Our method, despite being zero-shot on this dataset, has the best performance in 5 out of 6 configurations---all configurations except for the canonical orientation with top-k prompt searching. This searching process takes over an hour on an A100 and our method is not designed for test-time prompt searching.}
\label{table:miou_shapenetp_full}
\end{table*}

\begin{table*}[h!]
\centering
\begin{center}
\begin{small}
\begin{tabular}{p{0.12\textwidth}|p{1.4cm}p{1.4cm}|p{1.4cm}p{1.4cm}|p{1.4cm}p{1.4cm}|p{1.4cm}p{1.4cm}}
\toprule
   mIoU(\%) & \multicolumn{4}{c|}{Canonical Orientation} & \multicolumn{4}{c}{Rotated}\\
\midrule
    & \multicolumn{2}{c|}{Full} & \multicolumn{2}{c|}{Subset} & \multicolumn{2}{c|}{Full} & \multicolumn{2}{c}{Subset} \\
\midrule
&\raggedleft\{part\} of a \{object\} &\raggedleft \{part\} &\raggedleft \{part\} of a \{object\} &\raggedleft \{part\} &\raggedleft \{part\} of a \{object\} &\raggedleft \{part\} &\raggedleft \{part\} of a \{object\} &\raggedleft \{part\} \arraybackslash\\
\midrule
PointCLIPV2 &\raggedleft 11.619 &\raggedleft 9.647 &\raggedleft 11.275& \raggedleft9.700& \raggedleft10.943& \raggedleft10.261& \raggedleft10.317& \raggedleft10.216 \arraybackslash\\
PartSLIP++ & \centering -- & \centering --&\cellcolor[rgb]{0.8,0.8,0.8}\raggedleft 5.123 &\cellcolor[rgb]{0.8,0.8,0.8}\raggedleft \textbf{32.705} & \centering -- & \centering -- &\cellcolor[rgb]{0.8,0.8,0.8}\raggedleft 3.866 & \cellcolor[rgb]{0.8,0.8,0.8}\raggedleft\textbf{23.033} \arraybackslash\\
OpenMask3D & \centering -- & \centering -- &\raggedleft 12.538 &\raggedleft 11.242 &\centering -- & \centering -- &\raggedleft 11.933 &\raggedleft 11.673 \arraybackslash\\
\method (Ours) & \raggedleft\textbf{17.143} &\raggedleft \textbf{16.211}&\raggedleft \textbf{16.861}&\raggedleft 16.384&\raggedleft \textbf{17.703}&\raggedleft \textbf{16.819}&\raggedleft \textbf{17.620}&\raggedleft 17.164 \arraybackslash\\
\bottomrule
\end{tabular}
\end{small}
\end{center}
\caption{Detailed results on \partnete test set. \colorbox{gray!35}{Shaded} cells mean the method is trained on the same dataset (expected higher than white cells), and white cells mean zero-shot evaluation. Cells with ``-" denote that the method is too slow to be evaluated on the full test set. We evaluate with 2 types of prompts: ``\texttt{\{part\} of a \{object\}}" and ``\texttt{\{part\}}". PartSLIP++ achieves the highest performance with the ``\texttt{\{part\}}" prompts, yet the performance drops $84\%$ when we vary the query prompt. This dataset is more challenging for our method due to the sparsity of labels and the presence of small parts that are not geometrically or colorfully prominent (\eg, buttons on a surface with the same color). 
Nevertheless, our method is more robust to rotation and prompt variation, and clearly outperforms the other baselines not trained on this dataset.}
\label{table:miou_partnete_all}
\end{table*}

\FloatBarrier

\end{document}